\documentclass{article}

\usepackage{graphicx}%
\usepackage{multirow}%
\usepackage{amsmath,amssymb,amsfonts}%
\usepackage{amsthm}%
\usepackage{mathrsfs}%
\usepackage{xcolor}%
\usepackage{textcomp}%
\usepackage{manyfoot}%
\usepackage{booktabs}%
\usepackage{algorithm}%
\usepackage{algorithmicx}%
\usepackage{algpseudocode}%
\usepackage{listings}%
\usepackage{array}
\usepackage{caption}
\usepackage{pdflscape}
\usepackage[authoryear]{natbib}

\usepackage{url}
\usepackage{hyperref}  % must come after natbib
\hypersetup{
  colorlinks=true,
  citecolor=blue,
  linkcolor=blue,
  urlcolor=blue
}
\usepackage{subcaption}
\usepackage{threeparttable}
\usepackage{hyperref}
\usepackage[draft]{fixme}
\fxsetup{inline,nomargin,theme=color}

\usepackage[margin=1in]{geometry}

\usepackage{amsmath}
\DeclareMathOperator*{\argmax}{arg\,max}
\DeclareMathOperator*{\argmin}{arg\,min}

\newcommand*\diff{\mathop{}\!\mathrm{d}}
\newcommand{\SecureKL}{\textsc{SecureKL}}
\newcommand{\Do}{\mathcal{D}_o}
\newcommand{\Di}{\mathcal{D}_i}

\newcommand{\DT}{\mathcal{D}_T}
\newcommand{\AUC}{\mathrm{AUC}}
\newcommand{\AUCo}{\mathrm{AUC}_{o}}

\newcommand{\AUCT}{\mathrm{AUC}_T}
\newcommand{\priorkl}{\citet{shen2024data} }
\newcommand{\priorp}{\cite{shen2024data} }
\newcommand{\KLXY}{\mathrm{KL}_{\mathcal{X}\mathcal{Y}}}
\newcommand{\KLX}{\mathrm{KL}_\mathcal{X}}

\definecolor{todo-color}{HTML}{ff3232}

\let\cite\citep

\title{Privacy-Preserving Dataset Combination}
\author{Keren Fuentes$^{*1}$, Mimee Xu$^{*2}$, Irene Y. Chen$^{3,4}$}
\date{$^1$Independent Researcher, $^2$New York University, $^3$University of California, Berkeley, $^4$University of California, San Francisco \\ $^*$Equal contributions }

\begin{document}

\maketitle

\abstract{% % per AAAI instructions, no citations in abstract ~\cite{pmlr-v219-compton23a, shen2024data}
Privacy concerns and competitive interests impede data access for machine learning, due to the inability to \emph{privately} assess external data's utility. This dynamic disadvantages smaller organizations that lack resources to aggressively pursue data-sharing agreements. In data-limited scenarios, not all external data is beneficial, and collaborations suffer especially in heavily regulated domains: metrics that aim to assess external data given a source e.g., approximating their KL-divergence, require accessing samples from both entities pre-collaboration, hence violating privacy~\cite{pmlr-v219-compton23a, shen2024data}. This conundrum disempowers legitimate data-sharing, leading to a false ``privacy-utility trade-off".
  To resolve privacy and uncertainty tensions simultaneously, we introduce {\SecureKL}, the first secure protocol for dataset-to-dataset evaluations with zero privacy leakage, designed to be applied preceding data sharing. {\SecureKL} evaluates a source dataset against candidates, performing dataset divergence metrics internally with private computations, all without assuming downstream models. 
  On real-world data, {\SecureKL} achieves high consistency ($>90\%$ correlation with non-private counterparts) and successfully identifies beneficial data collaborations in highly-heterogeneous domains (ICU mortality prediction across hospitals and income prediction across states). Our results highlight that secure computation maximizes data utilization, outperforming privacy-agnostic utility assessments that leak information.}

\section{Introduction}\label{sec:intro}

\begin{figure}[t]
\centering
\includegraphics[width=0.80\linewidth]{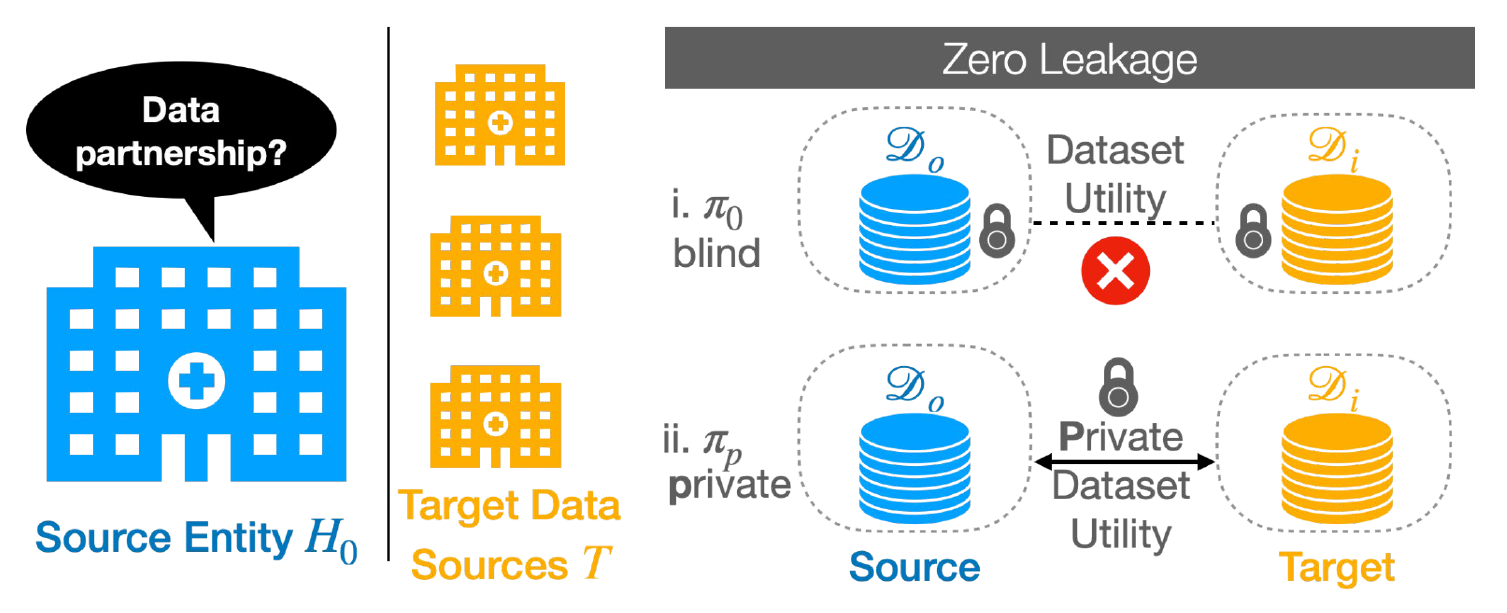}
        \caption{\textbf{Privacy can dis-incentivize data collaborations.}
        %When collaborations are resource-intensive, 
        %Lacking certainty in externally-held data, t
        Without seeing external data, an organization has two strategies:
        \textbf{i}. \textbf{blind default $\pi_0$}: randomly selecting partnerships causes hesitation and hinders partnerships. 
         \textbf{ii}. \textbf{private evaluation $\pi_p$}: securely assessing datasets \emph{before} commitment.}
    \label{fig:problem-cartoon}
\end{figure}

Since~\citet{Hestness2017DeepLS}, empirical works have predetermined data as a key driver to performance gains, via the so-called ``scaling laws''~\cite{Kaplan2020ScalingLF,brown2020language}. Yet, accessing and combining datasets is persistently challenging. As datasets have evolved from small to larger~\cite{sun2017revisiting}, more diverse~\cite{zhang2020pushing}, and more compute-optimal~\cite{hoffmann2022training}, the field of machine learning continues to seek more data ~\cite{mahajan2018exploring,brown2020language,park2019specaugment,reed2022generalist,sheller2020federated} and better ways to \emph{combine} it ~\cite{shen2023slimpajama, nguyen2023culturax,petty2024does,li2024datacomp, mckinney2020international}.

Strategically combining data from different sources promises enhanced models, but disempowers smaller organizations. While diverse, high quality data often improves performance, robustness, and fairness~\cite{miller2021accuracy}, access to such data significantly varies across entities and domains~\cite{tenopir2011data, 7906512}. 
As domain-specific data becomes increasingly valuable~\cite{alsentzer2019publicly,gururangan2020don,lee2020biobert}, data-owning entities are more reluctant to share it for free, opting instead to sell it in emerging data markets~\cite{acemoglu2022too,huang2021toward,liang2018survey}. This dynamic disproportionately handicaps smaller organizations who lack both the resources to purchase data and the leverage to negotiate favorable sharing agreements.

Organizations may hesitate to commit to a potential partnership when unsure about the benefits. 
As Figure~\ref{fig:problem-cartoon} illustrates, this ``commitment issue" is not solely a privacy issue; it's the inability to privately assess an external dataset's utility \emph{before} partnerships. This evaluation is, however, nontrivial. For example, adding in-domain data does not necessarily result in more performant models. Due to the precarious nature of domain shifts, machine learning models are sensitive to additional training data, often unpredictably when the source dataset is small (e.g., a single hospital's data) -- a phenomenon known as ``the dataset combination problem"~\cite{koh2021wilds,wang2022generalizing,taori2020measuring, bradley2021unrepresentative,meng2018statistical,shen2024data}.

\begin{figure}[t]
    \centering
    \begin{minipage}[t]{0.43\linewidth}
    \includegraphics[width=1.0\linewidth]{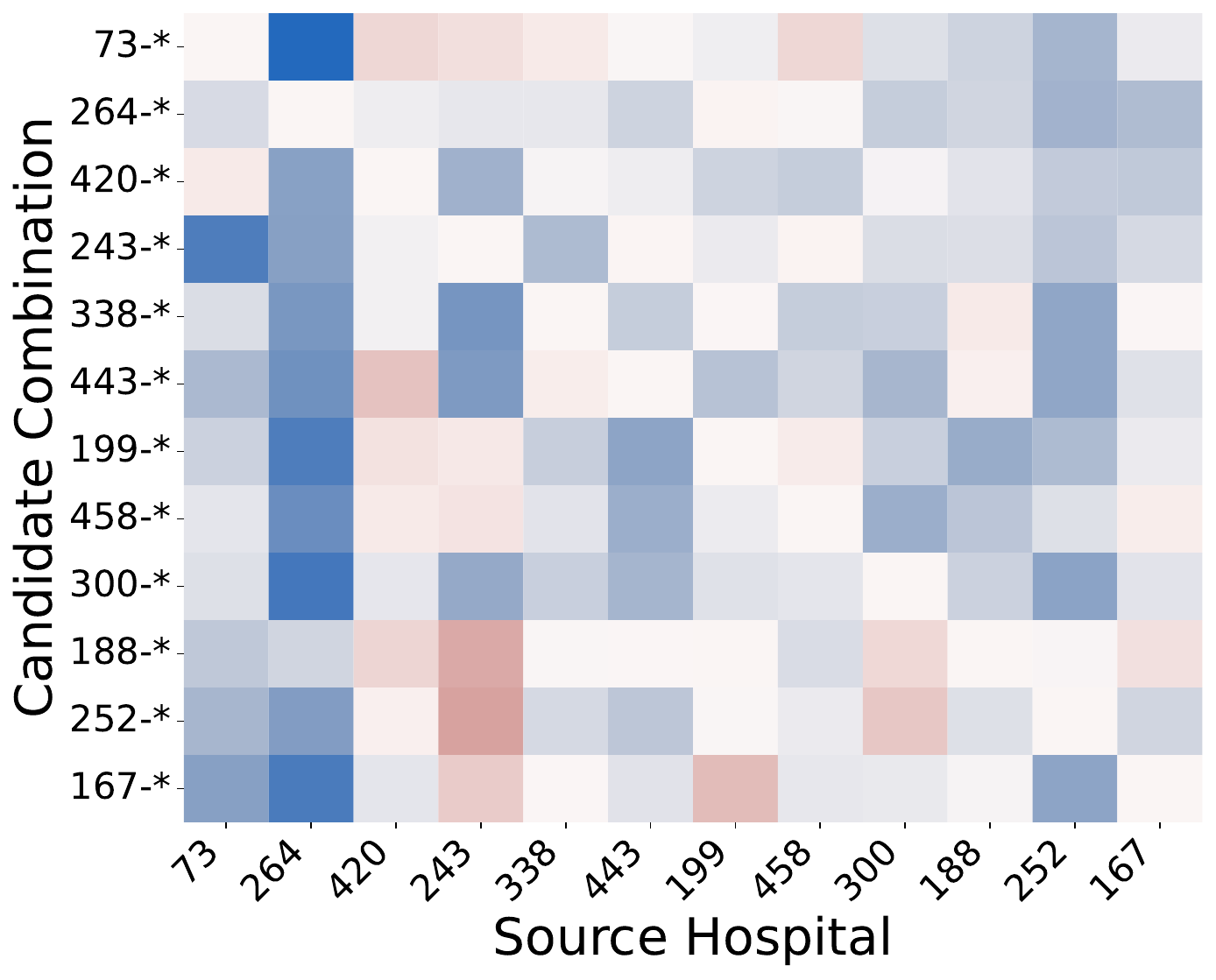}        
    \end{minipage}
    \hfill
    \begin{minipage}[t]{0.44\linewidth}
    \includegraphics[width=1.0\linewidth]{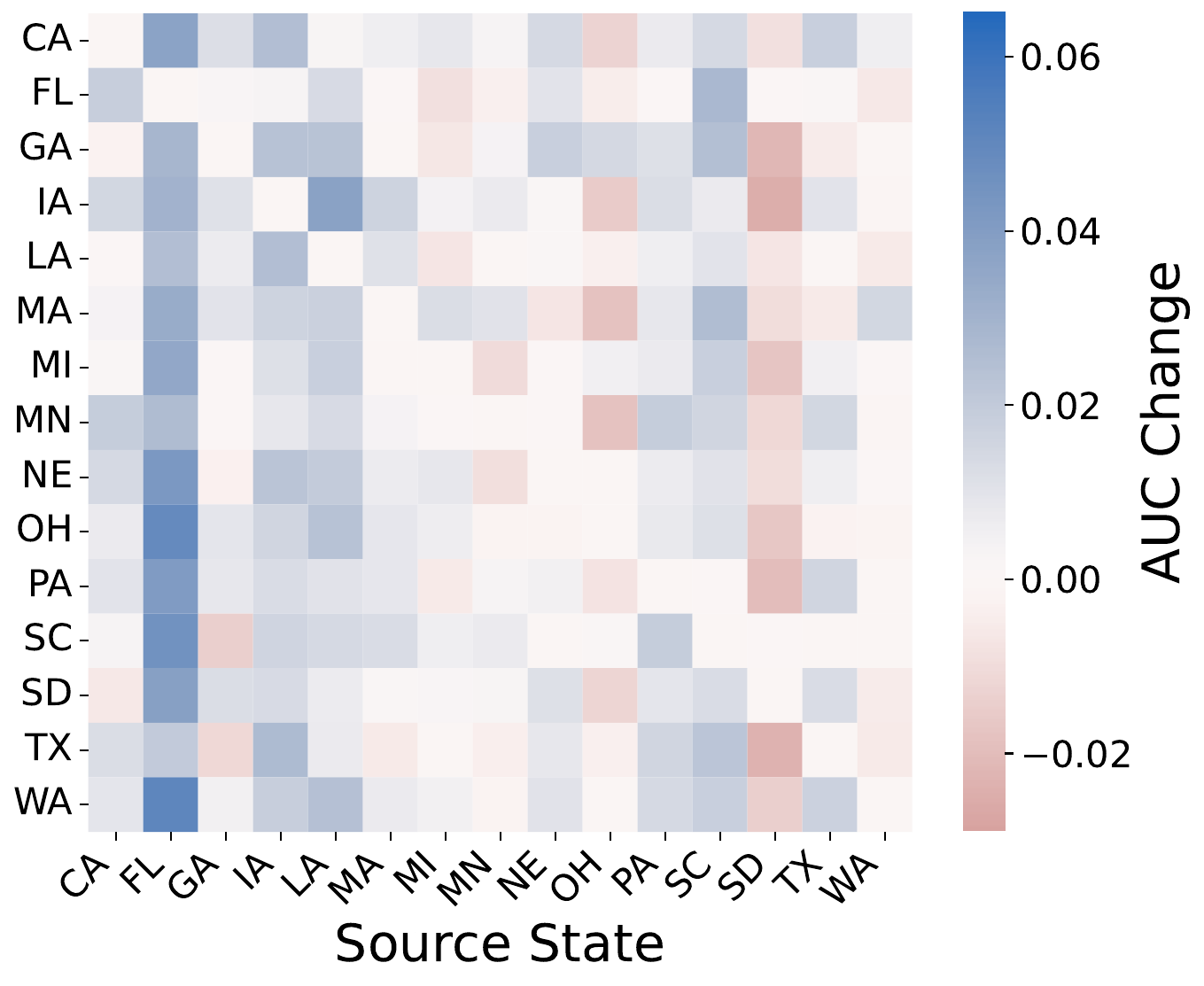}       
    \end{minipage}
    \caption{\textbf{The Dataset Combination Problem.} Real-world data collaborations are inherently uncertain. %necessitating pre-partnership selection.
    AUC change for a source entity, after incorporating external data across hospitals and states. 
    Each entry representas training on a pairwise data combination for a test entity (\textbf{x-axis}) 
    Each column represents training on a data combination for a fixed test dataset, formed by pairing the fixed entity with an external dataset in the y-axis. 
    \textbf{Left}: In eICU~\cite{Pollard2018TheEC},
    10 out of 12 hospitals may see their mortality prediction model degrade for \emph{some} potential hospital partners. \textbf{Right}: In Folktables~\cite{Ding2021RetiringAN}, combining with random state leads to worse income prediction in 10 out of 15 states. (red is bad; exact values are reported in Appendix~\ref{app:folktables})
    %The risks of ``bad data deals'' can thwart collaborations. In this scenario, $\mathrm{SKL}$ validates partnerships robustly and safely before any data is exchanged.
    }
    \label{fig:data_add}
\end{figure}

%Unfortunately, addressing these data availability issues demands prioritizing the most relevant potential partnerships, without seeing their data. In regulated domains like healthcare, data sharing isn't merely limited by privacy concerns; instead, the real barrier is the inability to privately assess external data's utility. This evaluation is shrouded in a web of complications. %This dynamic creates an expensive, opaque landscape that severely disadvantages smaller organizations lacking resources for aggressive data acquisition. 
%It is false, for example, that adding in-domain data necessarily results in more performant models. Due to the precarious nature of domain shifts, when the source dataset is small -- such as in a single hospital -- the downstream machine learning model is very sensitive to the additional training data, often in unpredictable ways --- a phenomenon known as ``the dataset combination problem"~\cite{koh2021wilds,wang2022generalizing,taori2020measuring, bradley2021unrepresentative,meng2018statistical,shen2024data}.

%to directly address the crux thwarting potential collaborations, 
%as the unpredictable outcome is a salient disincentive
%Preventing squandering on harmful collaborations necessitates pre-partnership selection.

Ideally, \emph{all} target data should be considered to reduce uncertainty in costly data collaborations. Yet, \textbf{datasets owned by separate entities cannot be directly and fully accessed}, significantly limiting the practicality of non-private dataset measures~\cite{shen2024data, ilyas2022datamodels}.

Our work directly addresses this crux by recognizing both privacy and competitive incentives. First, \emph{before} committing to acquiring unseen data, we enable organizations to privately gauge the relative utility of candidate datasets. Second, we provide strong privacy guarantees required of entities operating under stringent regulations e.g. healthcare providers to navigate data acquisition. 

Developers are often uncertain about the most effective model before more data becomes available. This renders a secure data appraisal stage by~\citet{xu2022data}, which requires model parameters, not applicable. In this opportunistic setting, we ask: 

\textbf{\emph{Can we ascertain the differential utility of prospective datasets, without knowing the final model?}}

%Good
We introduce {\SecureKL}, \textbf{the first divergence-based dataset measure with zero privacy leakage}. Our key insight is that private divergence computations (via secure multiparty computation (MPC)) are more data-efficient than sharing samples, thus achieving high performance. By preserving performance, {\SecureKL} presents a compelling guarantee: \textbf{for both parties, privacy is fully protected while data utilization is maximized}. %{\SecureKL} protects privacy and maximizes the utilization of the samples.

\paragraph{Contributions} A novel secure dataset-to-dataset evaluation protocol {\SecureKL} (SKL) that reduces uncertainty in data utility under limited data and budget, producing privacy-preserving measures while using the maximum available samples. SKL achieves a $>90\%$ correlation with privacy-violating counterparts across two real-world heterogeneous domains. Empirically, on ICU mortality prediction, SKL reliably chooses beneficial hospital(s) to partner with, outperforming data-leaking alternatives, including using demographic summaries or sharing data subsets.

% \paragraph{Contributions} A novel secure dataset-to-dataset evaluation protocol {\SecureKL} (SKL) that reduces uncertainty in data utility under limited data and budget, producing privacy-preserving measures while using the maximum available samples. SKL achieves a $>90\%$ correlation with privacy-violating counterparts across two real-world heterogeneous domains. In downstream tasks such as ICU mortality prediction, SKL reliably selects beneficial partners and outperforms privacy-leaking heuristics like demographic summaries and data subset sharing.

\paragraph{Impact} We provide a practical solution for organizations seeking to leverage collective data resources while maintaining privacy and competitive advantages. Our major advantage lies in reliability, especially when small organizations cannot afford to invest in detrimental partnerships. These results demonstrate the potential for wider data collaboration to advance machine learning applications in high-stakes domains while promoting more equitable access to data. 
Our code is available, and can be readily deployed to demonstrate potential data value preceding collaborations\footnote{https://github.com/kere-nel/secure-data-eval}.
%  A novel secure dataset-to-dataset evaluation protocol {\SecureKL} (SKL) that reduces uncertainty in data utility under limited data and budget, producing privacy-preserving measures using maximum available samples, by leveraging a logistic membership model trained in private. 
% When data is limited, model performance becomes highly sensitive to data input, rendering sample utilization paramount. Secure computation relaxes a crucial constraint of working with sparse and uncertain information by enabling dataset measurements with \emph{all} underlying data.

\section{Background}\label{sec:prelim}
\subsection{Dataset Combination Problem}
In high-stakes domains, incorporating additional datasets may \emph{degrade} the model. In healthcare scenarios, both \citet{pmlr-v219-compton23a} and \citet{shen2024data} showed that blindly acquiring new datasets can degrade model performance.  We extend these findings by replicating the effect under our own experimental setup. As shown in Figure~\ref{fig:data_add}, combining data from one hospital or state with an external source often leads to inconsistent, and sometimes negative, changes in model performance. %Indeed, in both highly heterogeneous domains, opportunistically acquiring \emph{unseen} datasets may even be harmful. 
This non-monotonic behavior highlights the need to evaluate data partners \emph{before} embarking on a full-fledged collaboration. 

%\subsection{Data-leaking Measures Are Inefficient}
\subsection{Trading Off Sample Utility For ``Privacy"}
\begin{figure}[h]
\centering
    \includegraphics[width=0.88\linewidth]{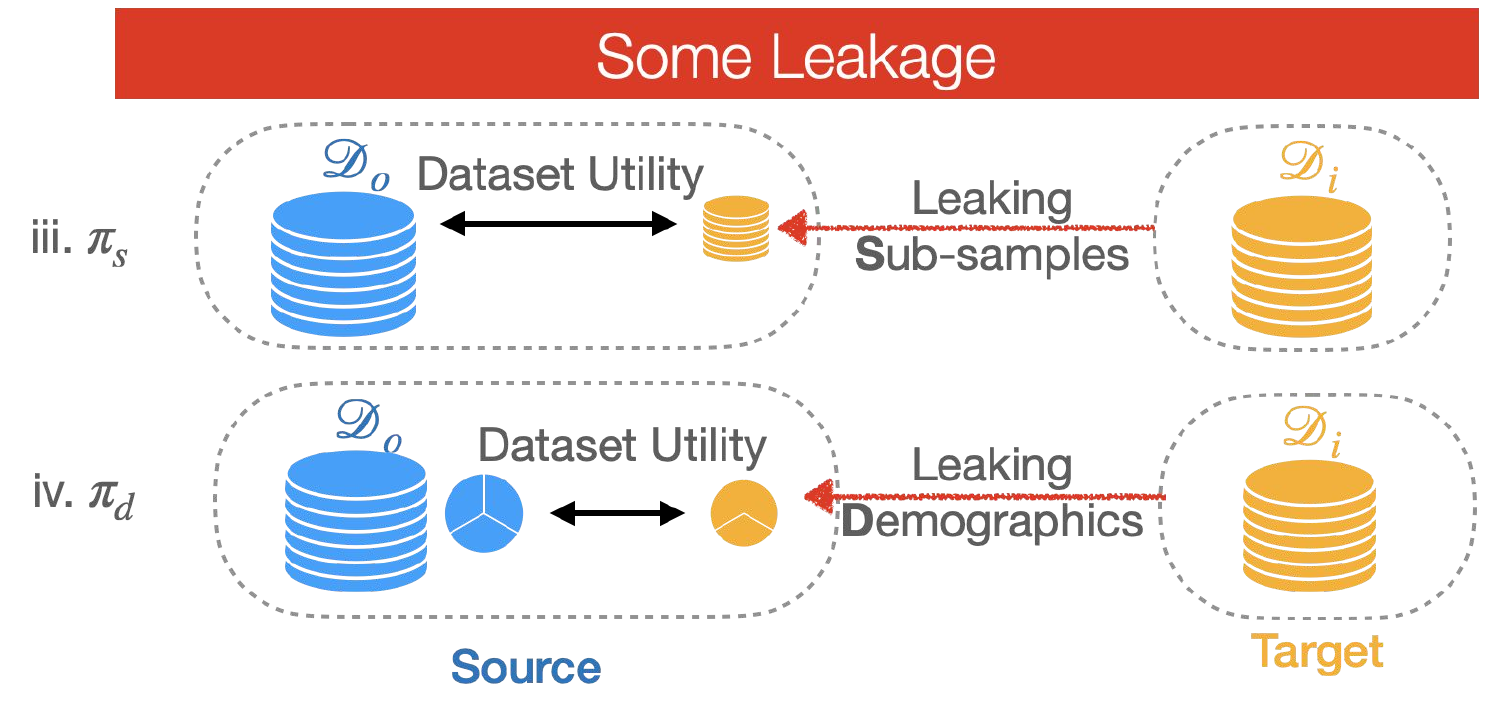}
    \caption{\textbf{Non-private evaluation strategies}.
        \textbf{iii}. \textbf{sub sampling $\pi_s$}: a subset of the target's data is shared. 
         \textbf{iv}. \textbf{demographic summaries $\pi_d$}: the target entity discloses distributions by protected attributes, i.e. age, gender, or race.}
    \label{fig:data-leaking-methods}
\end{figure}

Before an agreement, informed negotiations become impossible when entities do not expose data. In practice, data owners frequently resort to tiny samples~\cite{lds} or summary statistics (e.g., race, gender, age) for making decisions (Figure~\ref{fig:data-leaking-methods}). Yet, in data-limited settings, model performance is highly sensitive to new input. These heuristics are fickle, as sparse traits or limited samples often fail to capture the entire dataset's nuance and complexity, especially in heterogeneous domains, creating a perceived privacy-utility trade-off. We will show that secure computation, on the other hand, can avoid this tradeoff by measuring over \emph{entire} datasets while maintaining privacy.

% \paragraph{Challenges} We identify two unique conundrums in gauging unseen dataset's utility. One, \textbf{uncertain data utility deadlocks collaborations}. Computing utility \emph{after} data-sharing still disincentivizes partnerships. Federated secure model training, while privacy-preserving, still pre-supposes a collaboration before outputting each party's contribution~\cite{wang2020principled}, leading to a chicken-and-egg problem: the unpredictable outcome of partnerships in itself prevents partnerships from ever forming in the first place. Two, \textbf{poor sample utilization leads to privacy-utility trade-off}. Bounding data leakage naively handicaps utility computation, missing crucial details. For example, a hospital may choose another hospital with a similar gender ratio, avoiding collaborating with the rest, or they may ask for a tiny sample of data and extrapolate potential utility, yet such measures are noisy in heterogeneous domains, since few traits (or a few samples) rarely capture the wealth of a dataset. When most of the dataset information is foreclosed in the name of ``privacy", it is unclear whether those strategies can consistently proxy utility. When some leakage-bounded strategies are used to acquire data, privacy is sacrificed, but utility may still be limited. 

\section{Problem Setup}
\label{sec:problem}
Consider a binary prediction task for ICU patient mortality based on electronic medical records. A source hospital $H_o$ has historical patient data $\Do$ containing static past patient characteristics, prior medical records, and ICU outcomes. Other hospitals $\{H_i\}$ each has their patient data: $\{\Di \mid i\in [1.. N]\}$. 

For this binary prediction task, hospitals typically optimize for performance metrics, for example the area under the receiver-operating characteristic curve (AUC). Using only their data, $H_o$ can train a model $\mathcal{M}$ with parameters $\theta$ to achieve:
\begin{equation}
\tag{Baseline AUC}
\AUCo = \max_{f(\theta)} \, \AUC(\mathcal{M}, \Do)
\end{equation}
where $f$ is their chosen algorithm with parameter $\theta$.
%\footnote{\textcolor{red}{note that this algorithm can change downstream(for now we can omit)}}
%\AUC(H_o, \emptyset)
%, obtaining model parameters $\theta_o$, choosing an algorithm $f_o$, and an initial AUC of $\AUC^{[o]}$.

When $H_o$ has exhausted their own internal data, they may benefit from incorporating additional target data sources $T\subset [1.. N]$. By combining datasets, i.e., $\DT = \{\Di\mid i\in T\}\cup \Do$, $H_o$ can potentially achieve better results:
\begin{equation}
\tag{Combined AUC}
\AUCT = \max_{f(\theta)} \, \AUC(\mathcal{M}, \DT).
\end{equation}
We define the potential improvement from data addition as $\delta_{T} = \delta_{(o, T)} = \AUCT-\AUCo$. To add a single additional data source by setting $T=\{i\}$, the improvement is $\delta_{i} =\delta_{(o, i)}=\AUC_i-\AUCo$.
This leads to our central question:
\begin{quote}

    \textbf{\emph{Without seeing target data, how does a hospital ascertain potential data sources to combine with?}}
\end{quote}

Formally, given $n\leq N$, we seek a strategy $\pi$ that selects $n$ target datasets $T=\pi(\Do, n)$ to maximize model utility:
\begin{equation}
\tag{Ideal Combination}
\pi^*(\Do, n) = \argmax_{T\subset \binom{[1...N]}{n}}\AUCT%\quad\forall n.
\end{equation}
\paragraph{Practical Considerations.} Computing every subset $T\subset \binom{[1...N]}{n}$'s associated $\delta_{T}$ is exponential in $n$. To make this problem tractable, we make two key assumptions. First, we apply strategies greedily, selecting top-ranked target datasets. With the ultimate objective of improving the source hospital's prediction task, we fix $H_o$; to compare the trade-offs between strategies in Section~\ref{sec:methods}, we apply each $\pi$ greedily to select top-$n$ institution(s) for $H_o$ without replacement. Second, in data constrained settings, we aim to maximize the probability of positive improvement: $P_{H_o\sim \mathbf{H}}(\delta_T > 0)$. 
%\textcolor{red}{Add additional caveats here for folktable setups.}
\paragraph{Kullback–Leibler Divergence.} Our approach uses Kullback-Leibler (KL)-divergence-based methods to gauge data utility, building on prior work~\cite{shen2024data}. KL divergence~\cite{kullback1951information}, also called \emph{information gain}~\cite{quinlan1986induction}, describes a measure of how much a model probability distribution $Q$ is different from a true probability distribution $P$:
\begin{equation}
\tag{KL-Divergence}
\mathrm{KL}(P||Q) = \int_{x\in \mathcal{X}} \log\frac{P(\diff x)}{Q(\diff x)}P(\diff x)
\end{equation}
Because computing KL-divergence on hospital datasets $\Do$ and $\Di$ is non-trivial due to the high dimensionality of the data,~\protect\priorkl
proposes two groups of scores to make this divergence approximation tractable from small samples. We adopt one such score, $\KLXY$, which trains a classifier to distinguish between source and target samples and uses the classifier’s output probabilities to estimate dataset divergence.  We describe the $\KLXY$ score in detail in Section \ref{sec:KLXY}.

% \begin{equation}
% \tag{Ideal Estimator}
% \mathrm{KL}(P_o||P_i) = \int_{x\in \mathcal{X}} \log\frac{P_o(\diff x)}{P_i(\diff x)}P_i(\diff x)
% \end{equation}
 % Specifically, score $\KLXY$ first trains a logistic regression model on $\Do \cup \Di$ -- where the labels are folded into the covariates --- with the goal of inferring dataset membership. Then, the resulting model's probability score function $\text{Score}(\cdot): \mathcal{X, Y} \to [0,1]$ is averaged over a dataset in $H_o$, obtaining

% \begin{equation}
% \tag{$KL_{XY}$ Score}
% \KLXY = \mathbb{E}_{(x,y)\sim \Do}(\text{Score}(x, y)).
% \end{equation}
% \emph{Note}: This heuristic clearly hinges on model fit (over/underfitting can render the score less effective). Yet it reflects the insight that parametrized distribution estimations are more efficient on finite, \emph{unknown} data~\cite{nguyen2010estimating, valiant2017estimating}. We'll later show that $\KLXY$ is a reliable metric on real, heterogeneous data.

%Details are described in Section~\ref{sec:methods}.
\paragraph{Privacy Model} We operate under a semi-honest privacy model---also known as \emph{honest-but-curious} or \emph{passive security}---where parties follow protocols but may probe intermediate values. Parties are  ``curious'', meaning that they can probe into the intermediate values to avoid paying for the data. This assumes a weaker security model than malicious security where a corrupted party may input foul data, but ensures the algorithm to be private throughout the computation. This privacy preservation model incentivizes collaboration, improving upon~\protect\priorkl and ~\citet{ilyas2022datamodels}.

\paragraph{Dataset Divergence and Downstream Utility}
Building on the finding of \citet{miller2021accuracy} where a model's in-distribution performance is related to its out-of-distribution performance (across all model choices), we confer that dataset divergence does predict downstream model's performance after combining the data (Table~\ref{tab:encrypted_scores}), using metrics introduced in~\citet{shen2024data} (Section~\ref{sec:KLXY}). Intuitively, in-distribution quality is paramount in low-data settings, where dataset divergence can capture greater complexity and nuance than accessing a few traits.%\footnote{In contrast, data-rich domains like language modeling more frequently benefit from diverse, specialized data sources. 
Yet, privacy is unresolved: divergence measures entail accessing both entities' data~\cite{shen2024data,ilyas2022datamodels}, posing significant risks for heavily-regulated entities who are liable for any data exposure~\cite{hipaa, gervasi2022potential}.

%Competitive interests dictate that privacy and utility are not a ``balancing act" to trade off; ensuring privacy, such as through {\SecureKL}, is a necessity pre-partnership, even for small downstream improvements. 

%Incorporating information from entire datasets in a stable and robust metric demonstrates benefits, reduces uncertainty, and fosters trust before establishing collaborations. 
% Moreover, this data acquisition scenario needs to ensure privacy: the source data and the data to acquire are kept separate by default

% When data is limited, model performance becomes highly sensitive to data input, rendering sample utilization paramount. Secure computation, particularly \textbf{SecureKL}, relaxes a crucial constraint of working with sparse and uncertain information by enabling dataset measurements with \emph{all} underlying data.

%for both parties, \textbf{SecureKL} preserves privacy at the input and maximizes utility of the samples.
%unpredictable outcome is a salient disincentive
\paragraph{Secure Multiparty Computation (MPC)}
To cryptographically secure our divergence computation, we use Secure Multiparty Computation (MPC) ~\cite{yao1982protocols, shamir1979share}. MPC lets two or more parties to compute a function over private inputs, revealing only the final output~\cite{goldwasser2019probabilistic}.

% \textbf{Guarantee:} for both parties, privacy is guaranteed at the input, while samples are maximally utilized. 
Engineering machine learning workflows in private faces differs drastically from non-private machine learning engineering, due to 1. precision configurations and normalizations (to avoid blow-ups and underflows), 2. error control and performance engineering for machine learning in bit-limited spaces, and 3. debugging, hyperparameter tuning. Nevertheless, they are \emph{provably} secure.

Despite their non-trivial engineering, MPC programs enjoy strong security guarantees and relative ease of deployment. Even small organizations can deploy MPC without any specialized hardware. Thus, the algorithms developed and shared in {\SecureKL} readily enable zero-leakage dataset-to-dataset evaluations before sharing data. 

\paragraph{Assumptions}
% Assume:
% Source party has access to their own data (test set) where they want an algorithm to work well (though they may not know what algorithm model they use)
% Source party can buy / engage with data from other sources but they don’t have access directly
% (OR they only have a small percentage access)

% Goal: without compromising on data privacy, ascertain among candidate data sources, which ones would be sensible to combine with my setting and existing data?

% What is being done here?
Consider high stakes domains where disparate data may have additive benefits to the existing data. To make privacy boundaries tractable, we make the following assumptions:
\begin{enumerate}
\itemsep0em
\item \textbf{Existing knowledge} is not private. The hospitals are aware of each other having such data to begin with. The hospitals may know of the available underlying dataset size and format, which is assumed to be uniform across the hospitals in the setup to simulate unit-cost. Hospitals frequently know of each other's resources, and the available ICU units are contentious, not kept secret. 
\item \textbf{Uniformity} of $|\Di|$. Though each hospital gets to price their data and set their own budget, for generality, the uniformity assumption allows us to use the number of additional data sources $n$ as the main "budget proxy" across different strategies.
\item \textbf{Legal risks} of sharing \emph{any} data are omnipresent in high stakes domains. The risks with sharing sensitive data in data-leaking strategies, which we coin as $\pi_d$ (demographic distance) and $\pi_s$ (small sample), are not made explicit, but assumed to be "moderate" and "high" respectively. This abstraction side-steps legal discussion, which would go beyond the scope of our paper.
\item \textbf{No malice} is assumed on any of the parties involved, as each hospital wants to authentically sell their data and set up a potential collaboration. This assumption becomes stronger when the number of parties grows or when the setup changes to potentially more competitive industries with less trust. We note our limitations in Section~\ref{sec:limits}.
\end{enumerate}

\section{Methods}\label{sec:methods}
\label{eval_summary}
\begin{table}[h]
\centering

\begin{tabular}{p{0.25cm}lcl}
\toprule
 \multicolumn{2}{c}{\textbf{Strategy} }                     & \textbf{Sub-strategies} &\textbf{ Leakage} \\
\cmidrule(l){1-2}\cmidrule(l){3-3}\cmidrule(l){4-4}
$\pi_0$   & Blind {(baseline)}                         & n/a                 &   zero \\
$\pi_p$   & Private {({\SecureKL}) }                 & n/a                 &   minimal \\

$\pi_d$   & Demographic                    & {sex, gender, race}   &   moderate \\
$\pi_s$   & Subset Sampling                         & {1\%, 10\%, 100\%}     &  high\\
\bottomrule
\end{tabular}
\caption{\textbf{Concrete selection strategies $\pi$, differentiated by leakage risks}. For a strategy $p$, a set of targets is chosen, i.e., $T\gets \pi(\Do, n)$. %their risks proxy the potential cost of data acquisition. %Section~\ref{sec:strategies} describes the strategies.
    %, detailed in Section~\ref{sec:strategies}.
}
\label{tab:strategies}
\end{table}
\subsection{Defining Dataset Acquisition Strategies}
\label{sec:strategies}

Given the private and non-private data acquisition strategies illustrated in Figures~\ref{fig:problem-cartoon} and~\ref{fig:data-leaking-methods}, we associate privacy leakage as a primary cost of data partnerships.
This section formalizes them by their leakage risks, summarized in Table~\ref{tab:strategies}.% from all candidates. 
%This section formalize the corresponding strategies from Figures~\ref{fig:problem-cartoon} and~\ref{fig:data-leaking-methods}. Table~\ref{tab:strategies} associates them by three categories of data acquisition risks.
%As depicted by the strategies in Figures~\ref{fig:problem-cartoon} and~\ref{fig:data-leaking-methods}, the potential cost of data acquisition is linked to leakage risks. Therefore, we define three categories of risks and formalize their corresponding strategies, which are summarized in Table~\ref{tab:strategies}.

\textbf{A, high leakage}, sharing raw data. $\pi_s(n, k)$ supposes each hospital to share a dataset of size $k$; a default setting of $1\%$ is commonplace practice in some contracts, as a pre-requisite to being considered~\cite{lds}. Though leakage can be controlled through $k$, the data is inherently sensitive. The underlying distance metric follows the $\KLXY$ score introduced by~\protect\priorkl, which we define in the following subsection.

\textbf{B, moderate leakage}, sharing summary statistics. $\pi_d(n)$ uses demographic metadata to guide data selection. This is implemented through ratio distance between source and target distributions, which may be considered aggregates therefore potentially not sensitive, such as when the underlying aggregation function $\phi$ is differentially private.

\textbf{C, zero/minimal leakage}, sharing no \emph{additional} information besides what is assumed public, and what our method outputs, such as a score or a ranking. There are two methods: a. \textbf{Blind selection baseline}: $\pi_0(n)$ randomly selects $n$ disjoint data sources, until data purchasing budget runs out. Prior works suggests that when $n=1$, randomly selecting a source in hospital ICU may be risky and inefficient. b. \textbf{Our method} $\pi_p(n)$ selects data sources based on privacy-preserving measure for data combination, specifically Private $\KLXY$.

\subsection{KL-based Methods, Without Privacy}
\label{sec:KLXY}
In subset-sampling strategy $\pi_s(n, k)$, each of the candidate entities will leak a set of raw data. $\pi_s$ is implemented with KL-based measures similar to \citet{shen2024data}. Recall $\mathrm{KL}(P||Q)$ is not symmetrical, meaning that it is not a ``metric" that satisfies triangle inequality. Intuitively, this means the measure is directional: a hospital's distribution $P_o$ may be ``close" to the target distribution $P_i$, but not the other way around:
\begin{equation}
\tag{Ideal Estimator}
\mathrm{KL}(P_o||P_i) = \int_{x\in \mathcal{X}} \log\frac{P_o(\diff x)}{P_i(\diff x)}P_i(\diff x)
\end{equation}
Because we only have access to finite data $\Do$ and $\Di$, approximations are needed.
Typically, a learned model can capture distributional information, used to estimate continuous entropy.
Thus the joint distribution of features and labels from both the source and target are included, with the goal of deriving an efficient estimator for $\mathrm{KL}(P_o||P_i)$ that captures distributional shift from source to target.

Specifically, $\KLXY$ score used in $\mathrm{Secure}\KLXY$ first trains a logistic regression model \cite{Cox1958TheRA} on $\Do \cup \Di$ -- where the labels are folded into the covariates. The goal is to infer dataset membership using “proxy labels", defined as follows:  
$$\mathcal{I}(x,y) = \begin{cases}
        1 & \text{if } (x,y)\in \mathcal{D}_o \\
        0 & \text{if } (x,y) \in \mathcal{D}_i
\end{cases}$$
A binary predictor is fit on this combined dataset, predicting $\mathcal{I}$ from $(\mathcal{X}, \mathcal{Y})$ using logistic regression. The model's output $p(x,y)$, also called the probability score, is Score$(x,y)$.
% \begin{equation}
%     \tag{Logistic Regression}
%     \mathrm{LR}(x)
% \end{equation}
% We may wanna plug in the datasets in an equation here for clarity.

A score of $0.5$ or less means the datasets are not distinguishable, making the data potentially useful.~\protect\priorkl established the insight that in data-limited domains of heterogeneous data sources, domain shifts of the covariates are useful for predicting whether the additional data helps the original task, similar to~\citet{miller2021accuracy}.
We note again that even though this model is trained on both parties' data, the final algorithm that the hospital uses to train on combined data is not restricted.

Then, the resulting model's probability score function $\text{Score}(\cdot)$: $\mathcal{X, Y} \to [0,1]$ is averaged over a dataset in $H_o$, obtaining
% \begin{equation}
% \tag{KL-X}
% \KLX = \mathbb{E}_{(x)\sim \Do}(\text{Score}(x)).
% \end{equation} 
\begin{equation}
\tag{KL-XY}
\KLXY = \mathbb{E}_{(x,y)\sim \Do}(\text{Score}(x, y)).
\end{equation} 
We reproduce~\protect\priorkl's results that $\KLXY$ is predictive of downstream change in AUC in hospital setting (Table~\ref{tab:k_corr}).
Let the score function $g_\mathrm{KL}$ be the approximate of $ \mathrm{KL}(\Do||\Di)$. The strategy selects the most likely hospital with the closest distance under the measure:
\begin{equation}
\tag{KL-based Strategy, \emph{in plaintext}}
\pi_s (n=1, k=K) = \argmin _{i\in [1..N]} \, g_\mathrm{KL}(\Do,\Di).
\end{equation}

When only a subset is available, this function is adjusted by swapping $\Di$ for $\Di' \subseteq \Di$ where $|\Di'| = k$. We denote the full dataset size as $K=|\Di|$.

Though leakage can be controlled through $k$, yet the data is inherently sensitive. In ICU data, simulate that a default of $1\%$ is shared, so $k=3000\times 1\% = 30$, though we run experiments with $k\in\{3, 30, 300, 3000\}$.

\subsection{{\SecureKL}: Private KL-based Method}
\begin{figure}[t]
  \centering
  \includegraphics[width=0.98\linewidth]{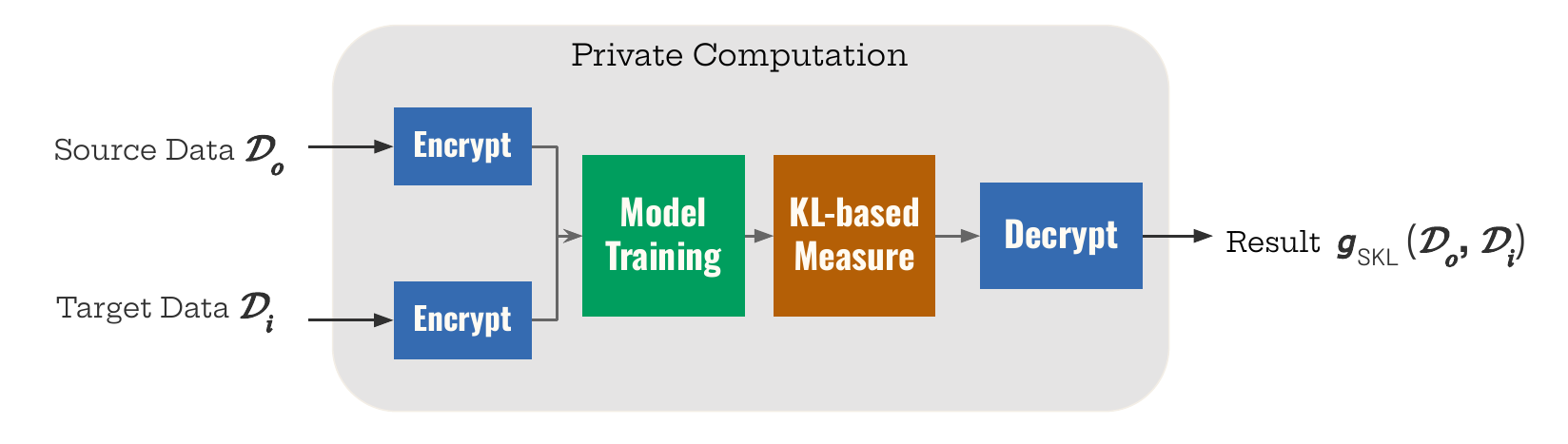}
  \caption{\textbf{Our method {\SecureKL}}. Each side encrypts their data. Then, a model is \emph{privately} trained on their joint data. Afterwards, their divergence is computed. Finally, only the final result of this dataset-to-dataset evaluation is revealed.}
  % \Description{Data is encrypted on each side.}
  \label{fig:method}
\end{figure}

Using MPC, we extend on $\KLXY$ to require no information sharing (besides what was already assumed public).
Specifically we leverage private tensors and secure gradient descent in CrypTen~\cite{knott2021crypten} to implement private $\KLXY$. As illustrated in Figure~\ref{fig:method}, the logistic regression as well as the scoring need to be implemented in private.
% Our code is publicly available \footnote{{https://github.com/kere-nel/secure-data-eval}}.

Denote the private encoding of $x$ as $[x]$.
% \begin{equation}
% \tag{Secure KL-X}
% \mathrm{Secure}\KLX = \mathbb{E}_{(x)\sim \Do}(\text{Score}([x])).
% \end{equation} 
\begin{equation}
\tag{Secure KL-XY}
\mathrm{Secure}\KLXY= \mathbb{E}_{(x,y)\sim \Do}(\text{Score}([x, y])).
\end{equation} 
Let the score function $g_\mathrm{SKL}$ be the secure approximation of $ \mathrm{KL}(\Do||\Di)$. The strategy selects the most likely hospital with the closest distance under the measure:
\begin{equation}
\tag{{\SecureKL} Strategy, \emph{encrypted}}
\pi_p (n=1) = \argmin _{i\in [1..N]} \, g_\mathrm{SKL}(\Do,\Di).
\end{equation}

As shown in Figure~\ref{fig:method}, any KL-based measure $g_\mathrm{SKL}$ can be adapted to our setup. We mainly use $\mathrm{Secure}\KLXY$ as the underlying measure. Its performance is detailed in Section~\ref{sec:positivity}. 
Additionally, even though our implementation measures distance of data between one source and one target party, the setup readily extends to accommodating multiple parties. Section~\ref{sec:eng-limits} discusses potential deployment challenges.

\begin{figure}[h]
  \centering
  \includegraphics[width=0.95\linewidth]{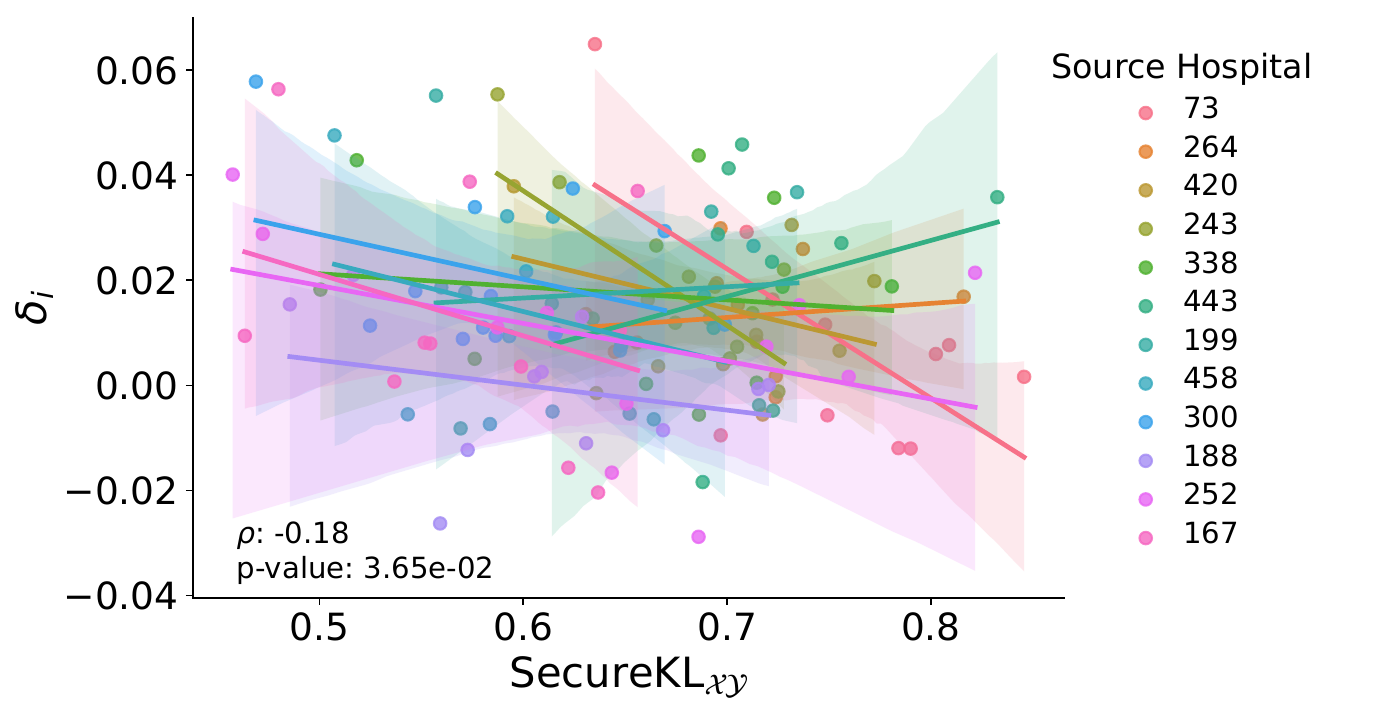}
  \caption{\textbf{{\SecureKL}: Overall Correctness}. Rank correlation between {\SecureKL} output and ground truth AUC change, $\delta_i$, from acquiring $1$ additional dataset for a given source hospital $H_o$. We propose selecting data partner ranked by our secure system under $\mathrm{Secure}\KLXY$ score to reliably increase AUC gains. ($|\mathbf{H}|= 12$ hospitals; colored by source.)}
  \label{fig:skl_delta}
\end{figure}

\subsection{Trivially ``Private'' Baseline: Blind Selection}
$\pi_0(n)$ randomly selects $n$ disjoint data sources, until data purchasing budget runs out. This random strategy may evade selection biases and help gather diverse data. Yet, prior work ~\priorp suggests that $\pi_0 (1)$ -- randomly selecting one source -- for ICU is risky and inefficient for mortality prediction.

\subsection{Alternative Data-leaking Baseline: Sharing Summary Statistics}
A relaxation to sharing no sensitive data is to share metadata. While demographic traits are often \emph{causal} and available, their exact cause in relation to the task is not a priori established (without a highly effective model), therefore their success in distributional-matching is not guaranteed to be strong. Additionally, in practice, the most effective model that results from data combination may or may not be causally-sound. Nevertheless, we posit alternative strategy
$\pi_d(n)$ to find the demographically close candidates to guide data selection: Let $\phi:\mathcal{D}\to\mathbb{R}^m$ be an $m$-dimensional summary statistic of a demographic feature i.e. the racial distribution of patients. Then, we use the distributional distance between $\Do$ and $\Di$, characterized by their $L_2$-distance through $\phi$:
\begin{equation}
\tag{Demographic-based Strategy}
\pi_d (n=1) = \argmin _{i\in [1..N]} \, L_2(\phi(\Do) || \phi(\Di)).
\end{equation}

\section{Experiments}\label{sec:exp}
\subsection{Experimental Questions}
\begin{enumerate}
\item \textbf{Consistency}: Does MPC degrade the original measure's effectiveness?
Since MPC implementations introduce approximations, we first validate $\mathrm{SKL}$'s correctness by examining the correlation of \textbf{private scores} ({\SecureKL}) vs. \textbf{plaintext scores} ($\KLXY$, with full data access). Then, for the consistency of choosing encryption in the hospital domains, we examine plaintext and encrypted versions' correlation with downstream ground truth rankings across hospitals.

\item \textbf{Positivity}:
Does our method pick entities that reliably improve performance? If source dataset $D_o$ can only add data from $n$ more sources, does our measure lead to eventual AUC improvements?
Specifically, in multi-dataset combination, we examine whether using $\mathrm{SKL}$ can improve the source hospital's downstream outcome. When selecting a single (or a few) additional data source, how many hospitals improve with our method? Additionally, we compare our method with alternative, privacy-leaking strategies.

\item \textbf{Error analysis}: If our privacy-preserving method is not the dominant strategy against alternatives including limited data accessibility, why?
As we know, small and uncertain improvements for downstream tasks underscore the inherent difficulty of evaluating data utility without seeing the full data. We are especially interested in analyzing (a) hospitals with low $\mathrm{Secure}\KLXY$ and $\KLXY$ correlations, and (b) hospitals lagging AUC improvements using the random strategy $\pi_0$ or limited-sample strategy $\pi_s$.
\end{enumerate}

\subsection{MPC implementation} $\mathrm{Secure}\KLXY$ includes training logistic regression model in private. We implement measures based on dataset divergence by building a custom logistic regression model over encrypted data leveraging CrypTen. The model parameters and input are encoded as 16-bit MPCTensors, ensuring that all computations, including forward passes, sigmoid activations, and gradient descent updates, are performed in private.

\paragraph{Additional baseline details}
We additionally run our experiments on plaintext methods used in~\citet{shen2024data}, including the $\KLX$ measure, which is similar to $\KLXY$ without using each data source's labels:
\begin{equation}
\tag{KL-X}
\KLX = \mathbb{E}_{(x)\sim \Do}(\text{Score}(x)).
\end{equation}
To compare against $\KLX$ for baseline, we additionally implemented its encrypted version, $\mathrm{Secure}\KLX$, as a $g_\mathrm{SKL}$ candidate.
\begin{equation}
\tag{Secure KL-X}
\mathrm{Secure}\KLX = \mathbb{E}_{(x)\sim \Do}(\text{Score}([x])).
\end{equation}

\paragraph{Optimizers} Because L-BFGS -- the optimizer prior work \priorp used in plaintext-only with Scikit-Learn~\cite{scikit-learn} -- is not available as an encrypted version, our MPC experiments are facilitated with SGD optimizer. A fair comparison between the scores obtained through plaintext and encrypted settings necessitates re-implementing plaintext scores, Score($X$) and Score($X,Y$), using logistic regression with SGD in PyTorch~\cite{paszke2019pytorch}. The hyper-parameter tuning for SGD in private and plain text are performed independently, as they do not transfer. Appendix~\ref{app:exp_details} will discuss hyperparameter-tuning in detail.

\subsection{Data and Model Setup}
\textbf{eICU dataset.} %Most abstract / highest level
The downstream task is $24$-hour mortality prediction from ICU data using the eICU Collaborative Research Dataset~\cite{Pollard2018TheEC}. This dataset contains $>200,000$ real-world admission records from $208$ hospitals across the United States.

\textbf{A note on data filtering.} Because medical research is inherently complex, ensuring reproducibility on statistical methods can be significantly challenging. ~\citet{Water2023YetAI} proposes that the research community follow a shared set of tasks with fixed preprocessing pipelines that are clinically informed -- essentially a machine learning training and evaluation protocol on eICU -- in order to facilitate method verification on the same benchmark. Therefore, our work follows their data cleaning criteria and the evaluation protocol. Additionally, we use the hospital exclusion criteria in \cite{shen2024data} to obtain top $12$ hospitals, where the most patient visits are collected (each with $>2000$ patients).

\textbf{Downstream model and baselines for eICU.}
% Middle
Recall that each strategy uses the same $K$ number of records per hospital -- in our experiment, $K=3000$. For $\pi_s$ which leaks a subset $k$ of all samples, a default $k = 1\% |\Di| $ randomly drawn samples are shared. In ICU data, we run experiments on $\{0.1\%, 1\%, 10\%, 100\%\}$.

Following holistic benchmarking tools in ~\cite{Water2023YetAI}, our strategy comparisons take $1500$ samples and the downstream model performance -- $\AUCo$, $\AUCT$ -- uses $400$ samples (unless otherwise noted).
% Details
Specifically, the AUC change, $\delta_i$ or $\delta_T$, comes from 1. combining $1500$ random samples from each selected dataset and 2. combining it with $1500$ samples from $\Do$, and 3. subtracting the baseline model's AUC\footnote{The samples are fixed across all experiments, the sample numbers are chosen to match \cite{shen2024data}'s setup.}.

\textbf{Folktables dataset} Though we primarily focus on hospital domain, we additionally validate using Folktables~\cite{Ding2021RetiringAN}, predicting across 15 states an individual’s annual income exceeds \$50,000\footnote{Because the Folktables dataset contains simpler features and is low-dimensional, KL divergence is directly computed rather than estimated. For implementation details, see Appendix \ref{app:folktables}.}. The details of our processing, which diverges from that of eICU, is included in Appendix~\ref{app:folktables}. 

\section{Results and Analysis}
\label{sec:eval_exp}
We organize our results around three research questions: whether using multiparty implementation sacrifices $\KLXY$'s efficacy (\textbf{consistency}; Section~\ref{sec:consistency}), whether our method reliably picks hospitals that improve performance (\textbf{positivity}; Section~\ref{sec:positivity}), and where our method may fail (\textbf{error analysis and limitations}; Section~\ref{sec:limits}).
\subsection{Consistency Between Plaintext and Encrypted Computations}
\label{sec:consistency}
%Our encrypted computations are programmed with CrypTen, when the plaintext counterpart is PyTorch-only.
Because our encrypted computation is the first implementation of dataset divergence in MPC, we ought to show that $\mathrm{Secure}\KLXY$ and $\mathrm{Secure}\KLX$ lead to highly comparable behaviors as $\KLXY$ and $\KLX$.

\paragraph{Spearman's Rank Correlation Coefficient for Underlying Scores}
For each source hospital $H_o$, use all full samples for $\Di$. Between $\KLXY$ and $\mathrm{Secure}\KLXY$ on $\Do$ and $\Di$ for all remaining hospitals $H_i$, $\mathbb{E}_{H_o\sim \mathbf{H}}[\rho] = 0.908$ with a range of $[0.691,1.0]$, obtaining $p < 0.02$ across all hospitals.
Between $\mathrm{Secure}\KLX$ and $\KLX$, $\mathbb{E}_{H_o\sim \mathbf{H}}[\rho]=0.9303$ with a range of $[0.455,0.991]$, with 11 of 12 hospitals achieving p-values below $0.05$. After applying Hochberg false discovery rate correction \cite{Benjamini1995ControllingTF}, our p-values remain significant. This range may be an artifact of sweeping hyperparameters independently in plaintext and encrypted optimisations, because sharing the same SGD hyperparameters would result in a tighter range. For all 12 hospitals, see appended Appendix~\ref{app:score_corr} for details.

\paragraph{Impact of Adding Security on AUC Correlation}
We further examine the effect by \emph{adding} encryption through its impact on the downstream AUC, using how AUC improvements are ranked. This rank is compared with how secure measures (i.e., $\mathrm{Secure}\KLXY$) and plaintext measures (i.e. $\KLXY$) rank hospitals. This comparison investigates the extent of the shift in the full hospital ranking, when we switch from a plaintext setup to encrypted. For $H_o\sim\mathbf{H}$, we measure $\delta_i$ that results from adding $\Di$ to $\Do$ for all $i$. This correlates all target hospitals $\{H_i\}$ with their ground truths $\{\delta_i\}$ in the case of picking a single target hospital.
We find the linear coefficient for encrypted $\mathrm{Secure}\KLXY$ to be $-0.182$ and plaintext $\KLXY$ to be $-0.184$ ($99\%$ matching). Both $\mathrm{Secure}\KLX$ and $\KLX$ have a linear coefficient of $-0.164$ with $\delta_i$.
For all strategies' correlations with ground truth at $n=1$, see Appendix~\ref{app:k_corr}.

\subsection{Positivity in Realistic Setup}
\label{sec:positivity}
\label{sec:exp_kl}
We apply {\SecureKL} in a pragmatic multi-source data combination problem, where each strategy acquires $n$ datasets for $n\in \{1,2,3\}$. 
\paragraph{Overall Positivity.} For $n=1$, we find that $\pi_p$ improves AUC in 10 out of the 12 hospitals. When $n=2$ and $n=3$, we find that using $\pi_p$ consistently improves AUC for all hospitals. Overall, 34 out of the 36 dataset combinations we evaluate on have an AUC improvement $\delta_T > 0$, suggesting that $\pi_p$ is a reasonable strategy for selecting hospital dataset combinations with a high expected return $\mathbb{E}[P_{H_o\sim \mathbf{H}}(\delta_T > 0)]$ for the source hospital from using our strategy.

\begin{table*}[t]\setlength\tabcolsep{3.1pt}
\centering
\begin{tabular}{clcccccc}
\toprule
& & \multicolumn{3}{c}{Demographic Summary Strategies} & \multicolumn{2}{c}{ Subset-sampling Strategies} & \multicolumn{1}{c}{Secure Strategy (\textbf{Ours})}\\ %(\textbf{Our Method})
\cmidrule(lr){3-5}\cmidrule(lr){6-7}\cmidrule(lr){8-8}
Dataset & $n$ & $\pi_d$-gender & $\pi_d$-age & $\pi_d$-race & $\pi_s (k=300)$ & $\pi_s (k=30)$ & $\pi_p\ \mathrm{Secure}\mathrm{KL}_{\mathcal{XY}}$ \\

\midrule
\midrule
eICU & 1 & $\textbf{0.020} \pm 0.023$ & $0.012 \pm 0.016$  & $0.014 \pm 0.015$ & $0.012 \pm 0.014$  & $0.010 \pm 0.017$ & $0.011 \pm 0.018$ \\
& 2 & $0.016 \pm 0.016$ & $0.017 \pm 0.017$  & $0.015 \pm 0.013$ & $0.024 \pm 0.020$ & $0.017 \pm 0.019$ & $\textbf{0.027} \pm 0.022$ \\
& 3 & $0.020 \pm 0.023$ & $0.016 \pm 0.019$ & $0.011 \pm 0.021$ & $0.021 \pm 0.017$ & $0.021 \pm 0.024$ & $\textbf{0.024} \pm 0.015$ \\
\midrule
Folktables & 1 & $0.006 \pm 0.017$ & $0.007 \pm 0.013$ & $0.006 \pm 0.009$ &  $0.006 \pm 0.016$ & $0.007 \pm 0.016$ & $\textbf{0.009} \pm 0.013$\\
            & 2 &  $\textbf{0.015} \pm 0.019$ & $\textbf{0.015} \pm 0.014$ &$0.010 \pm 0.013$ & $0.011 \pm 0.014$ & $0.014 \pm 0.018$ & $0.010 \pm 0.017$ \\
            & 3 &$0.013 \pm 0.018$
 & $0.014 \pm 0.016$ & $0.013 \pm 0.017$ &  $0.014 \pm 0.016$ & $\textbf{0.016} \pm 0.019$ & $0.011 \pm 0.017$ \\
\bottomrule
\end{tabular}
\caption{\textbf{AUC improvements in mean and standard deviation}, across all source regions for each strategy $\pi$, for eICU and Folktables setups. $n$ denotes the number of candidate datasets added to the source dataset. The small gains and high variance from adding selected datasets highlight the precarious nature of assessing data value in the real world. \textbf{Bold} indicates highest AUC improvement per $n$. \emph{Note}: Only $\pi_p\ \mathrm{Secure}\mathrm{KL}_{\mathcal{X}\mathcal{Y}}$ is private.}
\label{tab:combined_results}
\end{table*}

\begin{figure*}[t]
\begin{minipage}[t]{0.47\linewidth}
    % \begin{minipage}[t]{0.49\textwidth}
    % \includegraphics[width=0.98\linewidth]{figures/bp_demo1.pdf}
    % %\label{fig:demo_strats}
    % \end{minipage}
    % %
    % \begin{minipage}[t]{0.49\textwidth}
    \includegraphics[width=0.98\linewidth]{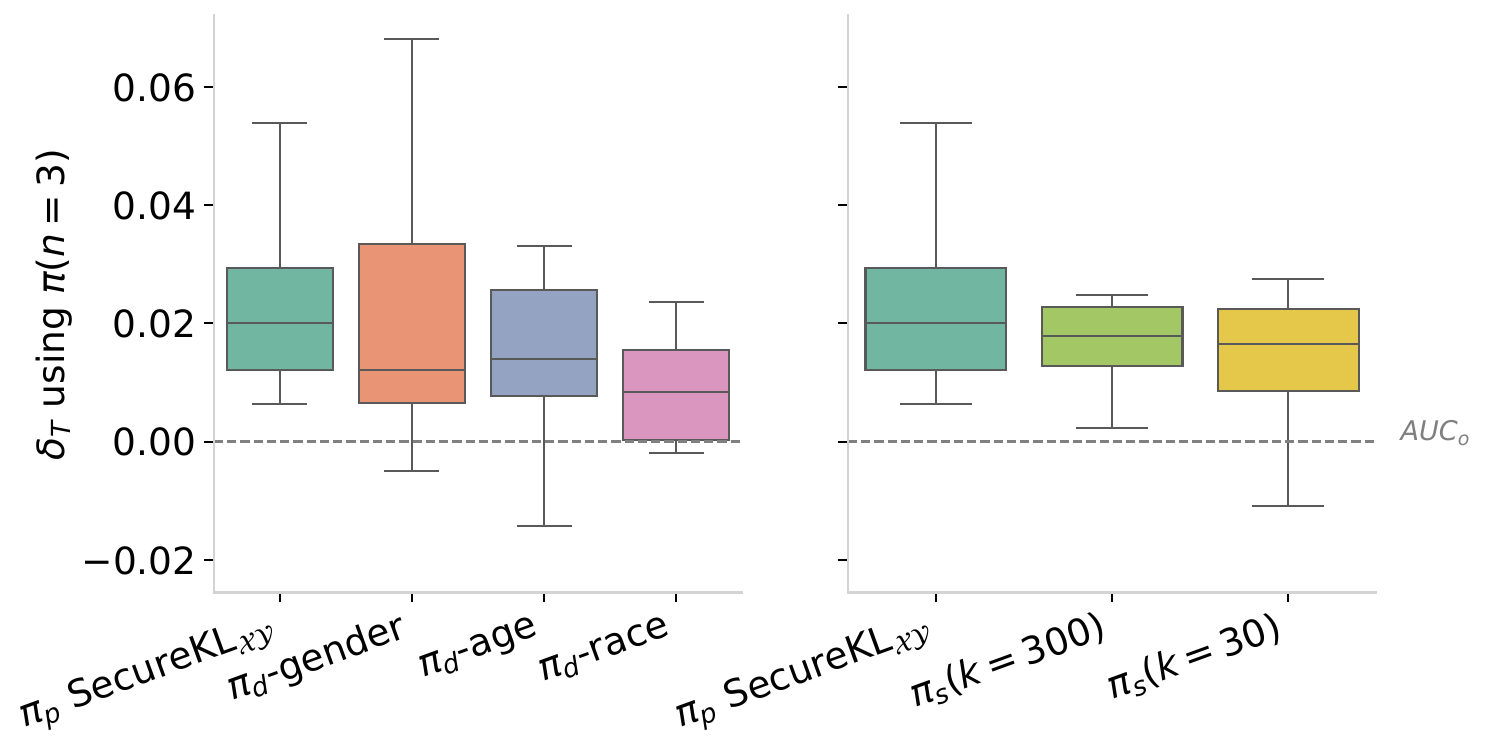}
    %\label{fig:p_strats}
    % \end{minipage}
      \caption{\textbf{AUC change $\delta_T$ over all strategies in eICU prediction} (higher is better). Our private dataset evaluation strategy $\pi_p$ outperforms demographic-based strategy $\pi_d$ \textbf{(left)}, and subset-sampling strategy $\pi_s$ for $k=300$ (10\%) and $k=30$ (1\%) \textbf{(right)}, after combining source data with the top $3$ candidates. }
      \label{fig:demo_strats}
    \label{fig:p_strats}
\end{minipage}
\hfill
\begin{minipage}[t]{0.47\linewidth}
    % \begin{minipage}[t]{0.49\textwidth}
    % \includegraphics[width=0.98\linewidth]{figures/folktables_bp_demo1.pdf}
    % \end{minipage}
    %
   % \begin{minipage}[t]{0.49\textwidth}
    \includegraphics[width=0.98\linewidth]{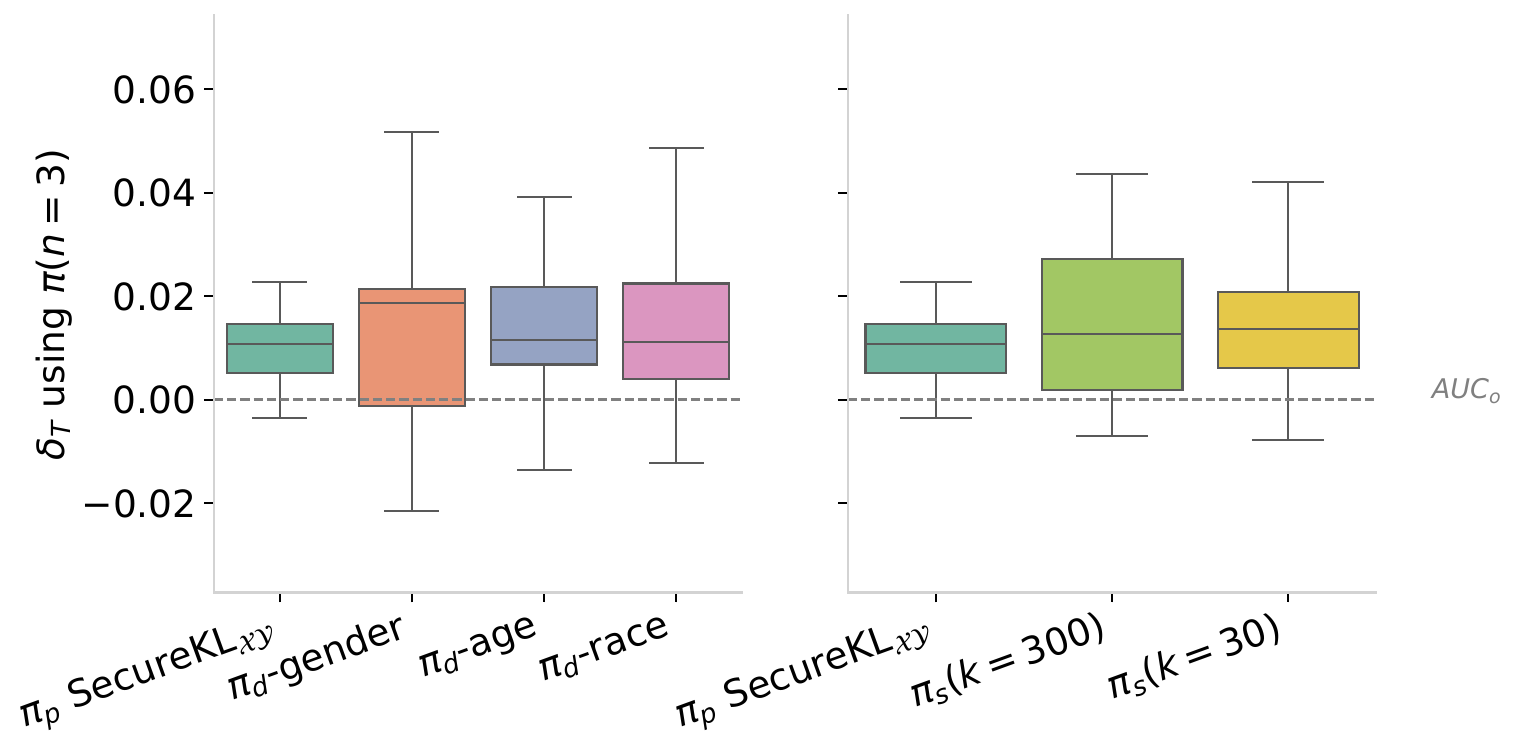}
    %\end{minipage}
      \caption{\textbf{AUC change $\delta_T$ over all strategies in Folktables dataset prediction} (higher is better). All strategies exhibit comparable distributions, after combining data from top 3 candidates. In a noisy domain, our method is stable: it neither excels nor penalizes against non-private strategies.}
      \label{fig:folktables_boxplots}
\end{minipage}
\end{figure*}
\paragraph{Comparing With Alternative Strategies}
\label{sec:strategies_discuss}
Other strategies -- $\pi_0$, $\pi_d$, and $\pi_s$ -- can also arrive at ``positive'' datasets. Comparing private method to other strategies at $n=3$, i.e, $\pi_p(n=3)$, we describe our results in Figure \ref{fig:p_strats}:
\begin{enumerate}
% a mean $\delta_T$ of $0.024$,, and a positive Bowley’s skewness of $0.085$.
\item $\pi_p$ (our method based on $\mathrm{Secure}\KLXY$) has a median $\delta_T$ of $0.020$, and a standard deviation of $0.015$. Our results indicate that for $50$\% of the hospitals, $\pi_p$ gives a $\delta_T >= .02$. Compared to other strategies, $\pi_p$ has the highest median, the lowest standard deviation, and it is one of two strategies that improves performance for all hospitals.

\item Demographic-based strategies underperform compared to $\pi_p$ on average. However, we observe that $\pi_d$-gender can be highly effective for a subset of hospitals, as it achieves the highest 75th percentile (Q3) of $0.033$ among all strategies. This indicates that for 25\% of hospitals, $\delta_T \geq 0.033$. Despite this, $\pi_d$-gender has a lower median value of $0.012$ compared to $\pi_p$, exhibits a high standard deviation ($0.022$), and degrades the performance for certain hospitals. Similarly, $\pi_d$-age has a median of $0.014$, and $\pi_d$-race has a median of $0.008$, both lower than $\pi_p$'s median.
% positive Bowley’s skewness of $0.583$.
% While $\pi_d$-gender a positive Bowley’s skewness of $0.583$.
% \item $\pi_d$-gender a median of $0.012$, and a Our results indicate that gender can be a great strategy for a subset of hospitals, as it has the highest maximum $\delta_T$ among all strategies and a positive skewness.
% However, it performs worse than $\pi_p$ on average, it has a high standard deviation ($0.022$), and it can degrade the performance of some hospitals.
%has a maximum $\delta_T$ of $0.068$, the trade-off lacks robustness, as this strategy has higher variability and degrades AUC a subset of hospitals.

% \item $\pi_d$-age has a minimum $\delta_T$ of $-0.026$ and a median $\delta_T$ of $0.014$. Compared to other strategies, $\pi_d$-age has the lowest minimum value.
% \item $\pi_d$-race has a median value of $0.008$. Our results indicate that for $50$\% of the hospitals, $\pi_d$ provides minimal improvements ($\delta_T < .008$).
\item Plaintext small-sample strategies, $\pi_s$, outperform all demographic-based methods but slightly underperform relative to $\pi_p$. For instance, $\pi_s(k=300)$ has a median $\delta_T$ of $0.0178$, and although it achieves $\delta_T > 0$ across all hospitals, it performs worse on average compared to $\pi_p$ and exhibits a higher standard deviation ($0.017$).
$\pi_s(k=30)$ has a median $\delta_T$ of $0.0165$. Compared to other strategies, it has the largest standard deviation ($0.024$), and it degrades the performance for some hospitals.
%$\pi_s(k=30)$ has a median $\delta_T$ of $0.0165$.

% \item $\pi_s(k=300)$ has a median $\delta_T$ of $0.0178$ and a standard deviation of $0.017$. Compared to $\pi_p$, this strategy also results in $\delta_T > 0$ for all hospitals, but it performs worse on average and it has a higher standard deviation.
% \item $\pi_s(k=30)$ has a median $\delta_T$ of $0.0165$, and a standard deviation of $0.024$. Compared to other strategies, it has the largest standard deviation, and it degrades the performance of a subset of hospitals.
\end{enumerate}

In summary, our method $\pi_p$ achieves the highest AUC improvement on average with the lowest standard deviation, demonstrating \textbf{more consistent improvements} for all hospitals. While the average improvement of $\pi_p$ is small, demographic-based and plaintext small-sample strategies exhibit greater variability, with some strategies improving performance for specific subsets of hospitals but underperforming or degrading results in others.
 % We visualize the impact of different strategies in Figure \ref{fig:demographic_strat} by measuring $\delta_i$. A positive $\delta$ means that a strategy $\pi$ improved the original performance of the model, while a negative $\delta_i$ means the AUC degraded after increasing the initial data fourfold.

 %We find that $\pi_p$ has a higher median and minimum $\delta$ than $\pi_d$, suggesting that $\pi_p$ not only performs better on average but also provides better worst-case performance, making it a more reliable strategy across hospitals. The gender-based $\pi_d$ strategy has a higher maximum $\delta$ for a subset of hospitals, indicating that it can result in substantial performance gains in certain cases. However, the trade-off is its lack of robustness, as it significantly degrades performance for other hospitals. Similarly, age-based and race-based strategies degrade AUC for a subset of the hospitals. While the $\pi_s(k=300)$ strategy performs similarly to $\pi_p$ on average, $\pi_p$ skews positively, indicating it tends to improve performance across most hospitals. In contrast, $\pi_s(k=30)$ skews negatively, suggesting it often results in worse performance, even below the original baseline, making it a less reliable strategy.
%While $\pi_s$ strategies have a similar AUC change as $\pi_p$ on average, $\pi_p$ skews positively and  $\pi_s(k=30)$ and $\pi_s(k=300)$ skews negatively, with $\pi_s(k=30)$ skew below the original baseline.

\subsection{SecureKL Error Analysis}
\label{sec:limits}

\paragraph{Underlying Data Limitations} In high-stakes domains, data partnerships are expensive, but potentially detrimental -- this forms a challenging landscape for evaluating methods on real-world data. Indeed, as shown in Table~\ref{tab:combined_results}, the AUC gain is small across all strategies, and the variance is high. This suggests that 3 hospitals' data is likely still too small for the general task to the robust explains limited AUC gains, highlighting the need to maximize samples for informative decisions. The key distinction, however, is that privacy-leaking methods (demographic, small sample) and blind baseline risk performance declines in many hospitals while SKL consistently improves downstream tasks \textbf{more reliably} than alternatives, across all hospitals, over multiple runs.

\begin{figure}[t]
\begin{minipage}[t]{0.46\textwidth}
  \centering
 \includegraphics[width=0.95\linewidth]{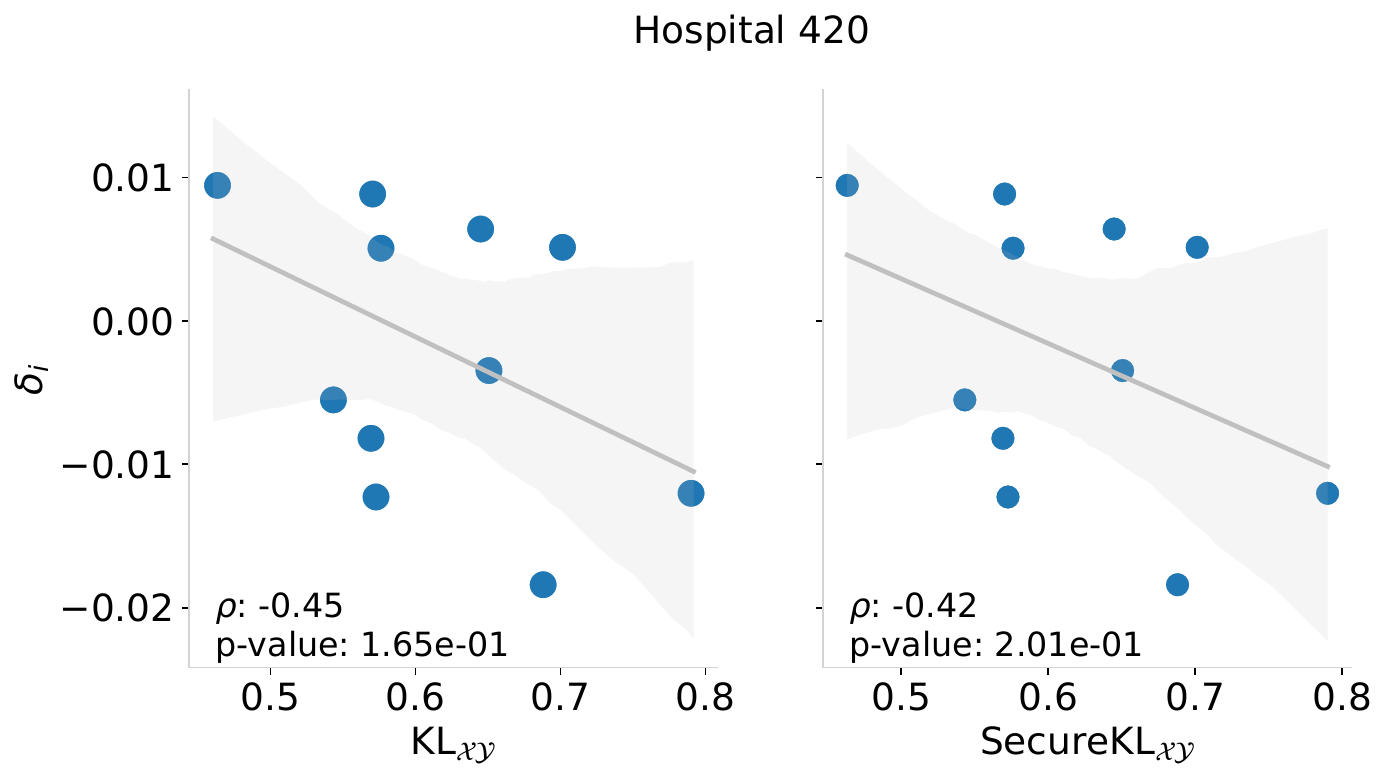}
 \end{minipage}
  \begin{minipage}[t]{0.46\textwidth}
  \centering
  \includegraphics[width=0.95\linewidth]{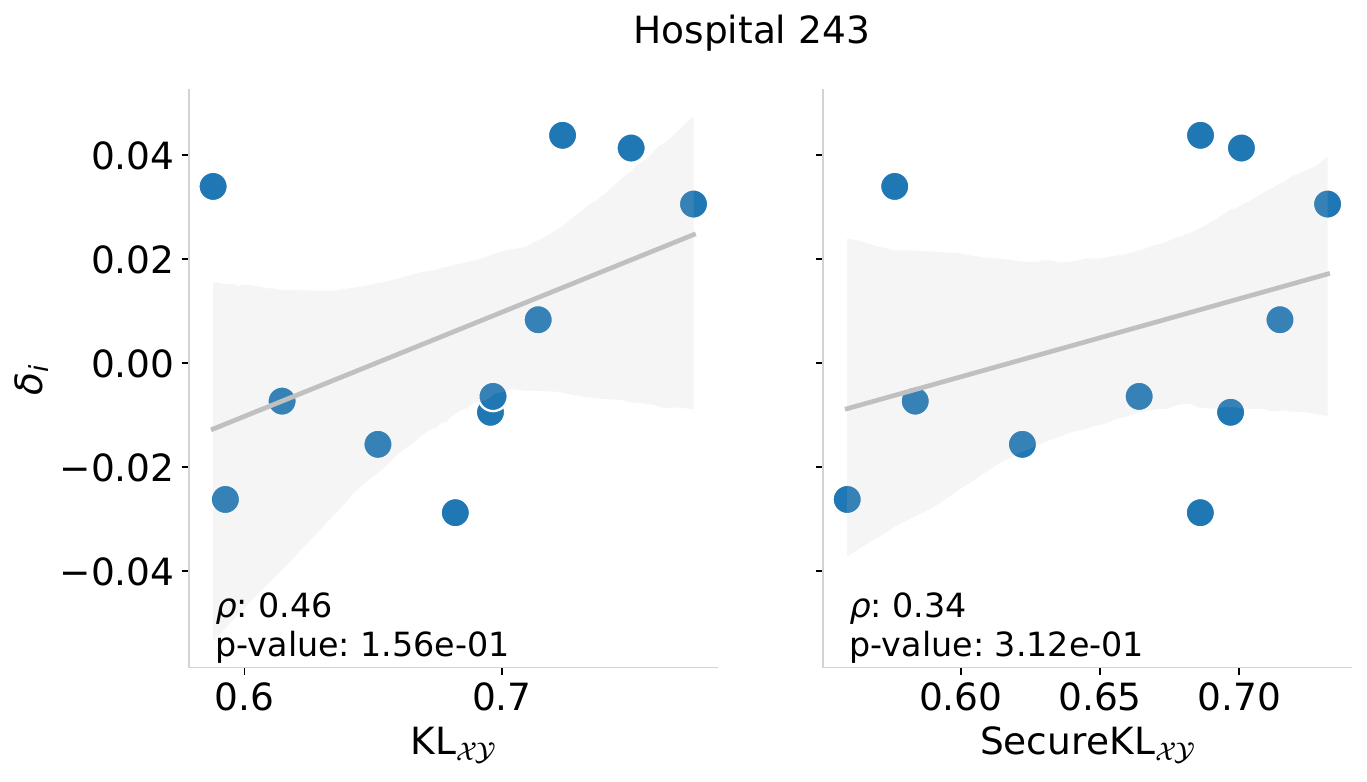}
  % \Description{\textcolor{red}{NEED UPDATING} }
  %\label{fig:corr_good}
\end{minipage}
  \caption{\textbf{The correlation between AUC change $\delta_i$ and underlying score $\KLXY$ affects $\mathrm{Secure}\KLXY$'s efficacy.} In Hospital 420, the $\KLXY$ scores identify beneficial data ($\rho<0$). For Hospital 243, $\KLXY$ anti-correlates ($\rho>0$). }
   \label{fig:hosp_corrs}
\end{figure}

\begin{figure}[t]
  \centering
  \includegraphics[width=0.99\linewidth]{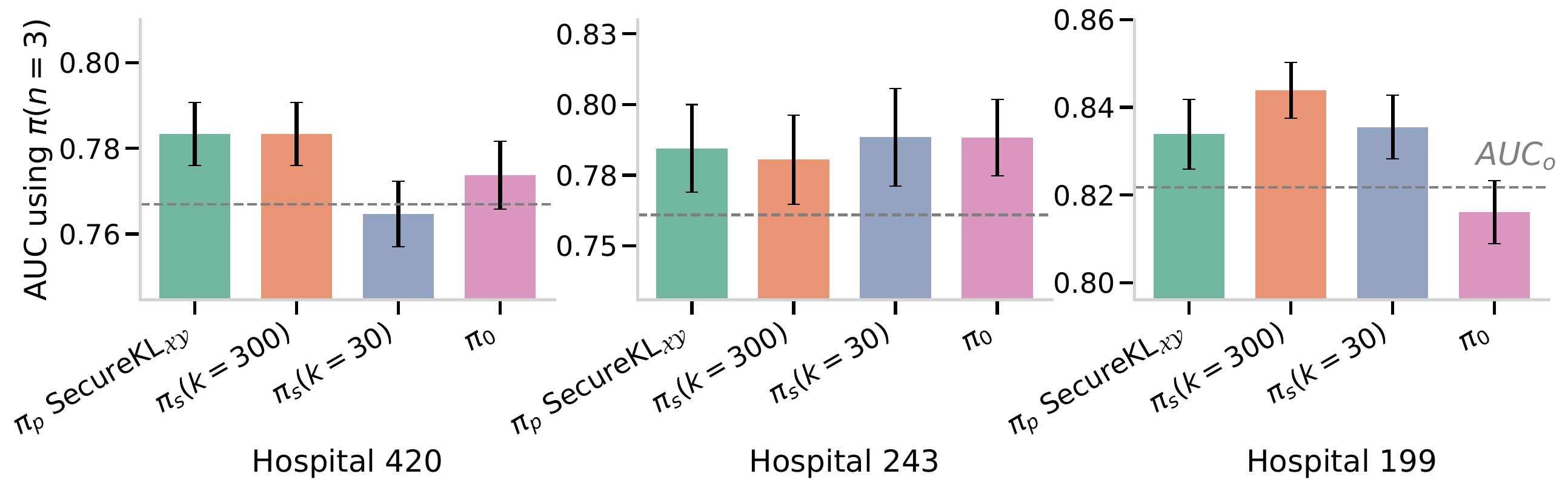}
  \caption{Mean AUC ($\pm$ standard deviation). \textbf{Left}: $\mathrm{Secure}\KLXY$ outperforms $\pi_s(k=30)$ and $\pi_0$. \textbf{Middle}: All strategies perform similarly. \textbf{Right}: $\pi_s(k=300)$ outperforms $\mathrm{Secure}\KLXY$.}  \label{fig:hos_auc}
\end{figure}

\paragraph{Underlying Score Limitations}
Data addition algorithms underpin the effectiveness of our method.
Even if $\Do$ obtains access to all the plaintext data, there is no guarantee that $\pi_p$ can correctly predict whether the data is useful. As seen in Figure~\ref{fig:hosp_corrs}, Hospital 243's utility when acquiring another data set is badly correlated with plaintext and encrypted KL-XY scores. This leads to its bad strategy for acquiring the top 3 hospitals, as seen in the middle pane of Figure~\ref{fig:hos_auc}. Interestingly, for this hospital, no other informational strategy excels, either, so choosing a random 3 may be preferred.

This behavior stems from the underlying measure, not from adding secure computation: in Figure~\ref{fig:hosp_corrs} and Figure~\ref{fig:hos_auc}, the encrypted performance closely follows that of plaintext performance, for both good and bad downstream correlations.

\paragraph{Sometimes, Not All Underlying Data Is Needed}
Relatedly, when seeing a few samples can successfully identify useful candidate hospitals, $\pi_s$ (which is on small samples) outperforms $\pi_p$ (which is on full samples).

In the right panel of Figure~\ref{fig:hos_auc}, hospital 199, the smaller sample sizes achieve a score that better reflects ground truth as a data addition strategy. In that case, the hospital may not need the full sample to know which target hospitals to collaborate with.

This behavior is specific to the interaction of the data and the underlying score, and does not affect the general insight that adding private computation preserves privacy (and eases privacy-related risks that hinder data sharing). We further note that our method still clearly applies to encrypted computation on a smaller data set under data minimization.

\section{Discussion}
\subsection{SecureKL Contribution}
\label{sec:skl_discuss}
After establishing that $\pi_p$ with $\mathrm{Secure}\KLXY$is a robust strategy in practical downstream performance, we hereby summarize the benefits of $\mathrm{SecureKL}$ and elaborate on their practical implications.

\textbf{Matching Plaintext Performance in Downstream Tasks.}
Our major contribution is to match plaintext performance with no data sharing. Using MPC provides \emph{input privacy}, meaning that if both hospitals only want to know the resulting score, the computation can be done without leaking original data. This strong guarantee can significantly ease the tension related to privacy and compliance in setting up a collaboration, leading to a practical "data appraisal stage" in data-limited high stakes domains.

\textbf{Gain from Data Availability.}
In contrast to limited-sample approaches, a key advantage for our method $\pi_p$ is that it takes advantage of all the underlying data -- generally impossible with non-secure methods for private data in heavily regulated domains. The general intuition is that data is localized; therefore, once a good target hospital is identified, we would prefer to acquire all of the data. It may be tempting to assert that we prefer the highest $k$ for data addition algorithms as well. In our experiments, while this is generally true, the smaller $k$ sometimes outperform larger $k$ in plaintext strategy $\pi_s$, which we investigate in Section~\ref{sec:limits} and in Figure~\ref{fig:hos_auc}. This occasionally non-monotonic behavior mirrors the challenge of data combination itself: even within one source dataset for the same estimator, more data is not necessarily better. This suggests the potential for a hospital-specific alternative to sharing a large amount of data for some source hospital, and points to future directions to using secure computation on a minimal-sized sample dataset for minimal performance overhead while remaining private.

\textbf{Potential Improvements to SecureKL} In the case where that output can be sensitive, i.e., when a source hospital queries a target hospital multiple times and accrues information through the score function, the \emph{output} can also be made privacy-preserving through differentially private data releases, such as using randomized response~\cite{dwork2014algorithmic}.

\subsection{Potential Challenges to Broader Adoption}
\label{sec:eng-limits}
Our code is readily usable by small organizations. While our approach generalizes to model-based measures (by substituting $g_\mathrm{SKL}$) and scales to multiple parties, our work also uncovered deployment limitations.

\begin{enumerate}
    \item \textbf{Operational}: engineering personnel limitations. While our implementation requires little cryptographic knowledge to deploy, it still needs technically-trained staff at each participating hospital to collaborate and maintain. This skill is similar to using pre-packaged software, cleaning data, and setting up network calls.

    \item \textbf{Engineering Extensions}: Extending any MPC protocol is non-trivial, as security engineering is a specialized skill. While $\mathrm{SecureKL}$ applies broadly to other underlying scores in multi-party setups, \emph{validating} a new MPC algorithm requires software engineering -- prototyping, tuning, debugging -- and numerical verification -- akin to data analytics and research - likely requiring technical talents who can be especially costly for hospitals to retain in-house.

    \item \textbf{Framework Limitation}: While CrypTen is designed to accommodate PyTorch, it is a research tool where not all plain text functionalities are implemented. For example, writing optimizers -- such as L-BFGS --  and custom operators that are not readily available requires both machine learning and cryptography knowledge. Moreover, the protocol incurs additional computational overhead, especially if hyper-parameters become more complex to sweep\footnote{For our work, the performance metrics are provided in Appendix~\ref{app:perf} for reference}. This will likely improve with time, as new frameworks address design shortcomings.

    \item \textbf{Inherent to Secure Computation}: When the method requires significant hyper-parameter tuning, such as using SGD on small batch data with learning rate schedules, plaintext tuning may not transfer perfectly. As detailed in Appendix~\ref{app:exp_details}, our hyperparameters for SGD differ in encrypted and plaintext settings. Thus, as encrypted computation \emph{hides} loss curves and training details by default, development is expected to be complex. This is because both hospitals want to ensure model fit with secure evaluation, but may not want to expend the computational cost of private hyperparameter sweeping.

\end{enumerate}
\section{Related Work}
\label{sec:related}
% 1. KL-divergence based methods are used for OOD detection, but not for acquiring new data. we follow up with prior work by adding a privacy mechanism (data addition).
%In addressing data access issues, we focus on incentivizing exchanges.
We briefly relate alternative approaches towards dataset privacy for sharing. More work on data pricing, differential privacy, and private fine-tuning are included in Appendix~\ref{app:extra_related}.
\paragraph{Augmenting existing data with synthetic data in medical domains}
Synthetic data generation has emerged as a promising approach to expand training datasets while preserving privacy. Generative adversarial networks (GANs) have shown success in generating realistic cancer incidence data~\cite{goncalves2020generation}, medical imaging data~\cite{thambawita2022singan}, and electronic health records~\cite{baowaly2019synthesizing}. These methods preserve statistical properties of the original data while providing differential privacy guarantees. Transforming data into a similar form that desensitizes certain attributes can be desirable~\cite{drechsler2011synthetic, howe2017synthetic, nikolenko2021synthetic, gonzales2023synthetic, sweeney2002k}. Yet, to still preserve the utility of the dataset transformed for analytics or learning tasks is challenging by itself~\cite{jordon2021hide}. Additionally, outside the scope of sensitive data that is transformed, little privacy guarantee is available, leading to re-identification risks~\cite{narayanan2006break, jordon2021hide}.

In addition, evaluation of synthetic medical data reveals challenges in capturing rare conditions and maintaining consistent relationships between multiple health variables~\cite{goncalves2020generation}. For tabular data, methods like CTGAN and TVAE~\cite{xu2019modeling} have demonstrated ability to learn complex distributions while preserving correlations between features. However, these approaches often struggle with high-dimensional data and can introduce subtle biases that impact downstream model performance~\cite{assefa2020generating}. Recent work has also explored combining synthetic data with differential privacy to provide formal privacy guarantees~\cite{jordon2018pate}. While these methods offer stronger privacy protection, they often face significant utility loss, particularly for rare but important cases in the original dataset~\cite{yang2024tabular}.

\paragraph{Secure Data Combination}
Recent work has explored methods for securely combining datasets while preserving privacy and improving model performance. Early approaches focused on using secure multi-party computation to enable multiple parties to jointly train models without sharing raw data~\cite{aono2017privacy}. However, these methods often struggled with computational overhead and communication costs when dealing with large-scale datasets~\cite{mcmahan2017communication}.
More recent techniques have introduced frameworks for evaluating potential data partnerships before commitment. These approaches use privacy-preserving protocols to estimate the compatibility and complementarity of different datasets~\cite{leung2019towards,chakraborty2024privacy}. Some methods focus specifically on measuring distribution shifts between datasets without revealing sensitive information~\cite{duan2021flexible}. Others trained the downstream model in private, but limited to LASSO~\cite{van2021privacy}. Several systems have been developed to facilitate secure data combination in specific domains. In healthcare, methods have been proposed for securely combining patient records across institutions while maintaining HIPAA compliance~\cite{raisaro2018m, van2021privacy}. Financial institutions have explored similar approaches for combining transaction data while preserving client confidentiality~\cite{liu2021finbert}.

% The field has also seen development of protocols for dynamic data partnerships, where entities can continuously evaluate and adjust their data sharing relationships ["AdaptiveShare: Dynamic Privacy-Preserving Data Integration Framework" by Lee et al., 2022; "SecureFlow: Privacy-Preserving Data Exchange Pipeline" by Martinez et al., 2023]. These methods often incorporate techniques from both secure computation and differential privacy to provide formal guarantees while maintaining utility ["PrivatePool: Scalable Privacy-Preserving Data Pooling" by Singh et al., 2022].

\paragraph{Federated Learning}
Cross-silo federated, decentralized, and collaborative ML~\cite{mcmahan2017communication, li2020federated, bonawitztowards, kairouz2021advances} focus on acquiring more data through improved data governance and efficient system design. Healthcare machine learning is considered especially suitable, as health records are often isolated~\cite{rieke2020future,xu2021federated, nguyen2022federated, cho2025secure}. Yet, even though no raw data is shared, model parameters or gradients flow through the system. As the federated computing paradigm offers no privacy guarantee, the system is vulnerable to model inversion~\cite{geiping2020inverting} and gradients leakage attacks~\cite{boenisch2023curious, zhu2019deep}. A subtle but urgent concern is that privacy risks discourage the very formation of the federation when optimization is traded off with privacy~\cite{lyu2022privacy,raynal2024conflict}.  %This is partly due to placing heavy trust on a party that can observe shared data such as model updates and gradient information [8-9. trusting central]. This setup may risk inherent optimization loss that directly trades off privacy protection [robustness-privacy trade-off - 10]. 
Building on the insight that useful data is often disparately-owned, we tackle the specific incentive problem between pairs of players where one side trains the model, instead of scaling up the \emph{number} of parties through a federation. %In addressing data access issues, we focus on incentivizing exchanges. %We thus focus on incentivizing this exchange. %efficient, accurate, and private.

Compared to vanilla Federated Learning, an MPC system \cite{shamir1979share, yao1982protocols, bonawitz2017practical, knott2021crypten} provides stronger guarantee in terms of input security. Model owners and data owners can potentially federate their proprietary data, including model weights, training, and testing data, can work together under stringent privacy requirements. Our work extends the line of work by ~\cite{xu2022data, yang2024fedfed, bonawitz2017practical} that demonstrates the potential of incorporating MPC in various federated scenarios. On the practical side, unlike mobile-based networks for secure federated learning protocols~\cite{bonawitztowards}, our system assumes a smaller number of participants, where communication cost and runtime are not dominant concerns. 

\paragraph{Comparing with Federated Learning with Privacy Guarantees} Privacy incentivizes federation~\cite{usynin2024incentivising}. Specifically, preserving privacy between the parties under federated learning uses secure computation methods~\cite{truex2019hybrid, bonawitz2017practical, stripelis2021secure,cho2025secure, froelicher2021truly} combined with differential privacy~\cite{xu2023federated, usynin2024incentivising, avent2017blender}, trusted execution environments~\cite{mo2021ppfl} -- an approach that is coined "Privacy-in-Depth" in~\citet{kairouz2021advances}. 

Performing dataset evaluation with MPC, as in {\SecureKL}, can be seen as an extension of data federation, by adding a separate privacy-preserving component \emph{before} all parties commit to an entire federated learning system. It is low-commitment, as privacy is preserved by default, and there is no requirement to continue in any system; it tackles the incentives' problem blocking collaboration. By demonstrating utility, the two parties foster trust. Our work complements the line of ambitious systems that incorporate MPC in potential federated scenarios where the participants are semi-honest~\cite{yang2024fedfed, bonawitz2017practical, cho2025secure}.

% A differentially private algorithm preserves privacy even if the output is shared.
% Not input-privacy preserving -> Does not solve the problem that the parties don’t want to share data to begin with.
% Furthermore, DP-SGD does not use all of the data, which makes optimizing top performance challenging, especially in data-constrained setups %~\cite{feldman2020neural} suggest that a large model relies on duplicate data to learn, and ignores data that is ``long-tail'' in terms of frequency

% Sharing data with differential privacy preserves privacy to a certain extent, but has significant drawback in performance in machine learning contexts~\cite{abadi2016deep}.
 
\section{Conclusion}\label{sec:conclusion}
%  We are replicating prior work closely, additionally we are comparing plain- text, but not the full plaintext (accomplish what has been done before). The KL-score that is based on the fraction of the samples, the correlation is not strong (so we really need the full data, **use L-BFGS), blindly choosing and using demographic-only data -- race distribution.

% 1. privacy is valuable, makes transactions easier to make, cost-effective for making better models that result in better healthcare, builds trust. (privacy being beneficial). 
% 2. health outcome is improved. Consequential things that can benefit from having more data. Reduce errors by X percent if you use our method to sequentially select hospitals to work with.
Our work demonstrates that privacy-preserving data valuation can help organizations identify beneficial data partnerships while maintaining data sovereignty. Through {\SecureKL}, we show that entities can make informed decisions about data sharing without compromising privacy or requiring complete dataset access. As the AI community continues to grapple with data access challenges, particularly in regulated domains like healthcare, methods that balance privacy and utility will become increasingly critical for responsible advancement of the field. As noted in Section~\ref{sec:limits}, our approach has several limitations, including the fact that, despite impressive aggregate results, our method is less effective for individual hospitals, which motivates future work. %Additionally, our work presents opportunities for follow-up research.
Our method assumes static datasets and may not generalize well to scenarios where data distributions evolve rapidly over time. A sequential version of our framework may more closely model dynamic data collaborations. Future work should explore extending these techniques to handle more complex data types and dynamic distribution shifts while maintaining strong privacy guarantees.
\clearpage
% Our method advances prior work in several key ways. Beyond replicating existing KL-score computations in a privacy-preserving manner, we demonstrate that using complete datasets rather than sample-based estimates is crucial for accurate partnership evaluation - particularly since correlation strength diminishes with reduced data sampling. Our empirical results show that privacy-preserving data partnerships can deliver tangible benefits: hospitals using our method to select collaboration partners saw a X\% reduction in mortality prediction errors compared to demographic-only or random selection approaches. This improvement highlights how privacy-preserving methods can facilitate valuable data partnerships while building trust and maintaining data sovereignty, ultimately leading to better healthcare outcomes through more robust models.

\clearpage
\bibliographystyle{plainnat}
\bibliography{main}

\clearpage
\section{Methodological Details}
\textbf{Datasets} We use two datasets: 1. eICU Collaborative Research Dataset~\cite{Pollard2018TheEC} contains over $200,000$ admissions from $208$ hospitals across the United States. Following the data cleaning and exclusion criteria outlined by \cite{Water2023YetAI} and \cite{shen2024data},
We select $12$ hospitals with the highest number of patient visits (each with at least $2000$ patients) as our entire set of hospitals $\mathbf{H}$. Each strategy would compute with a max $K=3000$ records, as the total available data per hospital.

2. To evaluate broader applicability, we replicate a portion of our experiments on the 
Folktables \cite{Ding2021RetiringAN} dataset on income prediction is additionally used, which provides rich demographic and socioeconomic information on individuals across U.S. states. We predict whether an individual’s annual income exceeds 50,000.

\textbf{Data Treatment}
For each strategy, the same records available per hospital are used, with $K=3000$. 
 Performance -- $\AUCo$, $\AUCT$ -- uses $400$ samples (unless otherwise noted) \footnote{This follows training and evaluation protocols in Yet Another ICU Benchmark \cite{Water2023YetAI}.} The AUC change, $\delta_i$ or $\delta_T$, comes from 1. combining $1500$ random samples from each selected dataset and 2. combine it with $1500$ samples from $\Do$, and 3. subtracting the baseline model's AUC\footnote{The samples are fixed across all experiments, the sample numbers are chosen to match the setup in ~\priorp.}.
The downstream task is the $24$-hour mortality prediction.Strategy comparisons take $1500$ samples.
We simulate the problem setup for each hospital with the $24$-hour mortality prediction task.
Unless otherwise specified, all experiments follow the training and evaluation protocol in Yet Another ICU Benchmark \cite{Water2023YetAI}, using $1,500$ training samples and $400$ test samples per hospital. For the data combination experiments that compute AUC change $\delta_i$ or $\delta_T$, to match ~\priorp, we take $1500$ random samples from each selected dataset and combine it with $1500$ samples from $\Do$. To match ~\priorp, each hospital experiment was carried out using 5-fold cross-validation, repeated 5 times with different random seeds. AUC results are averaged first across folds, then across repetitions.

The strategy comparisons described are implemented using $1500$ samples for our training set and $400$ samples for our test set per hospital for all of our experiments unless otherwise noted. This follows training and evaluation protocols in Yet Another ICU Benchmark \cite{Water2023YetAI}.

\section{Correlation with downstream performance}\label{app:k_corr}
On Table \ref{tab:k_corr}, we report the Pearson correlations between $\pi_s(k=K)$ for $k \in \{3,30,300, 3000\}$ and $\delta_i$. 
On Table \ref{tab:encrypted_scores}, we report the Pearson correlations between different strategies and $\delta_i$.

\begin{table}[t]
\centering
\begin{tabular}{llcc}
\toprule
\textbf{Dataset} & \textbf{Data Addition $\pi$} & \textbf{Pearson $r$} & \textbf{p-value} \\
\midrule
\multirow{7}{*}{eICU} 
  & $\mathrm{Secure}\KLXY$ & -0.182 & \textbf{3.65e-02} \\
  & $\mathrm{Secure}\KLX$  & -0.162 & 6.27e-02 \\
  & $\KLXY$                & -0.184 & \textbf{3.47e-02} \\
  & $\KLX$                 & -0.162 & 7.13e-02 \\
  & Gender                 &  0.097 & 2.65e-01 \\
  & Race                   &  0.018 & 8.29e-01 \\
  & Age                    &  0.053 & 5.33e-01 \\
\midrule
\multirow{7}{*}{Folktables} 
  & $\mathrm{Secure}\KLXY$ & -0.196 & \textbf{3.63e-03} \\
  & $\mathrm{Secure}\KLX$  & 0.17  & \textbf{8.40e-03} \\
  & $\KLXY$                &-0.166& \textbf{1.37e-02} \\
  & $\KLX$                 & 0.099  & 1.44e-01  \\
  & Gender                 &  0.138 & \textbf{4.14e-02} \\
  & Race                   &  0.133 & \textbf{4.84e-02} \\
  & Age                    &  0.067  & 3.20e-01 \\
\bottomrule
\end{tabular}
\caption{Pearson correlation ($r$) between each data addition strategy $\pi$ and AUC drop, reported separately for the eICU and Folktables datasets. Statistically significant $p$-values $(p < 0.05)$ are bolded.}
\label{tab:encrypted_scores}
\end{table}

\begin{table}[ht]
\centering
\begin{tabular}{l ll @{\hskip 1cm} ll}
\toprule
& \multicolumn{2}{c}{\textbf{SGD}} & \multicolumn{2}{c}{\textbf{LBFGS}} \\
\cmidrule(r){2-3} \cmidrule(l){4-5}
$k$ & \textbf{$\rho$} & \textbf{p-value} & \textbf{$\rho$} & \textbf{p-value} \\
\midrule
3    &  -0.063 & 4.70e-01  &  -0.158 & 7.02e-02 \\
30   &  -0.082 & 3.47e-01  &   0.167 & 5.60e-02 \\
300  &  -0.059 & 5.00e-01  &  -0.097 & 2.70e-01 \\
3000 &  -0.184 & \textbf{3.47e-02} &  -0.284 & \textbf{9.47e-04} \\
\bottomrule
\end{tabular}
\caption{$\rho$ and p-value between AUC drop and plaintext KL using $k$ samples with SGD and LBFGS.}
\label{tab:k_corr}
\end{table}
\section{Hyperparamter Tuning}\label{app:exp_details}
We obtain Score(X,Y) by training a Logistic Regression model using SGD. We find that SGD requires hyper-parameter tuning in order to perform well when evaluated on Brier Score Loss. We used Optuna to perform hyper-parameters search.

\begin{table}[h]
\centering
\begin{tabular}{@{}lcc@{}}
\toprule
\textbf{Hyperparameter} & \textbf{Plaintext} & \textbf{Encrypted (MPC)} \\
\midrule
Learning rate           & $0.0795$              & $0.0974$ \\
Patience (epochs)       & $2$                   & $5$ \\
Tolerance               & $0.000117$            & $0.000132$ \\
Momentum                & $0.886$               & $0.907$ \\
Weight decay            & $1.81\times 10^{-9}$  & $8.14\times 10^{-7}$ \\
Dampening               & $0.0545$              & $0.0545$ \\
\bottomrule
\end{tabular}
\caption{SGD hyperparameters selected by Optuna for the plaintext and encrypted logistic regression models (objective: Brier score).}
\label{tab:sgd_hparams}
\end{table}
%ESCORES_DIR = "max_it1713_eta00.09740654837869905_alpha8.149790303222343e-07_tol0.0001326143453600182_pat5_mom0.9072287159978029_damp0.014369120736053117"

\section{Correlations between Encrypted Scores and Plaintext Scores}\label{app:score_corr}

\begin{table*}[t!]
\centering
\begin{tabular}{lcl cl}
\toprule
Hospital & $\rho(\KLX, \mathrm{Secure}\KLX)$ & p-value & $\rho(\KLXY, \mathrm{Secure}\KLXY)$ &  p-value  \\ \midrule
73 &
0.945 & 1.118e-05 &
1.000 & 0.0\\
264 &
0.973 & 5.142e-07 &
0.945 & 1.118e-05 \\
420 &
0.982 & 8.403e-08 &
0.991 & 3.763e-09 \\
243 &
0.973 & 5.142e-07 &
0.909 & 1.056e-04 \\
338 &
0.973 & 5.142e-07 &
0.982 & 8.403e-08 \\
443 &
0.964 & 1.852e-06 &
0.882 & 3.302e-04 \\
199 &
0.991 & 3.763e-09 &
0.973 & 5.142e-07 \\
458 &
0.873 & 4.546e-04 &
0.964 & 1.852e-06 \\
300 &
0.455 & 1.601e-01 &
0.691 & 1.857e-02 \\
188 &
0.718 & 1.280e-02 &
0.864 & 6.117e-04 \\
252 &
0.873 & 4.546e-04 &
0.809 & 2.559e-03 \\
167 &
0.764 & 6.233e-03 &
0.891 & 2.335e-04 \\
\bottomrule
\end{tabular}
\caption{Spearman Correlations $\rho$ for encrypted (in CrypTen) and plaintext (in PyTorch) KL-based methods} 
\label{tab:encrypted_scores_corr_with_plaintext}
\end{table*}
On Table \ref{tab:encrypted_scores_corr_with_plaintext}, we measure the Spearman correlations between $\KLX$ and $\mathrm{Secure}\KLX$, and between $\KLXY$ and $\mathrm{Secure}\KLXY$ for all hospitals. We find that all hospitals have statistically significant correlations with the exception of hospital 300's $\rho$($\KLX$,$\mathrm{Secure}\KLX$)
\section{Performance}
\label{app:perf}
Computing $\mathrm{Secure}\KLXY$ scores for 144 hospital pairs, each with at most 3000 samples, took 317 seconds, which is ~6.6X longer than $\KLXY$.
\section{Folktables Experiments}
\label{app:folktables}
\paragraph{Dataset.} We use the Folktables dataset, a benchmark derived from the U.S. Census American Community Survey (ACS), which provides rich demographic and socioeconomic information on individuals across U.S. states. We focus on the 2014 \emph{ACSIncome} task, which classifies whether an individual's annual income exceeds \$50,000 based on $10$ features. Our core experiments include the following states: \texttt{SD}, \texttt{NE}, \texttt{IA}, \texttt{MN}, \texttt{OH}, \texttt{PA}, \texttt{MI}, \texttt{TX}, \texttt{LA}, \texttt{GA}, \texttt{FL}, \texttt{CA}, \texttt{SC}, \texttt{WA}, \texttt{MA}.

\paragraph{Baseline model.}
For each state, we sample $3,000$ training samples and $400$ test samples. Unless otherwise noted, these splits are held fixed across all strategies and experiments. For each state, we train an XGBoost classifier on its $3000$ training examples and evaluate on its 400-example test set.

\paragraph{Data Combination Protocol.} 
For the data combination experiments, we combine the $3000$ samples from each selected dataset to combine with the $3000$ samples in $D_o$. 
\paragraph{KL divergence estimation.}
For eICU data,  we used  $\text{Score}(\cdot)$ to estimate data density due to its high dimensionality. In contrast, Folktables has only 10 features, so we follow~\priorkl: we apply normalized PCA, retain the top three principal components, then fit a Gaussian kernel density estimator (bandwidth selected via cross-validated grid search). Using these densities, we compute KL divergence in both plaintext and our privacy-preserving setting to align with the rest of our experiments.

\begin{figure*}[t]
    \centering
    \includegraphics[width=0.98\linewidth]{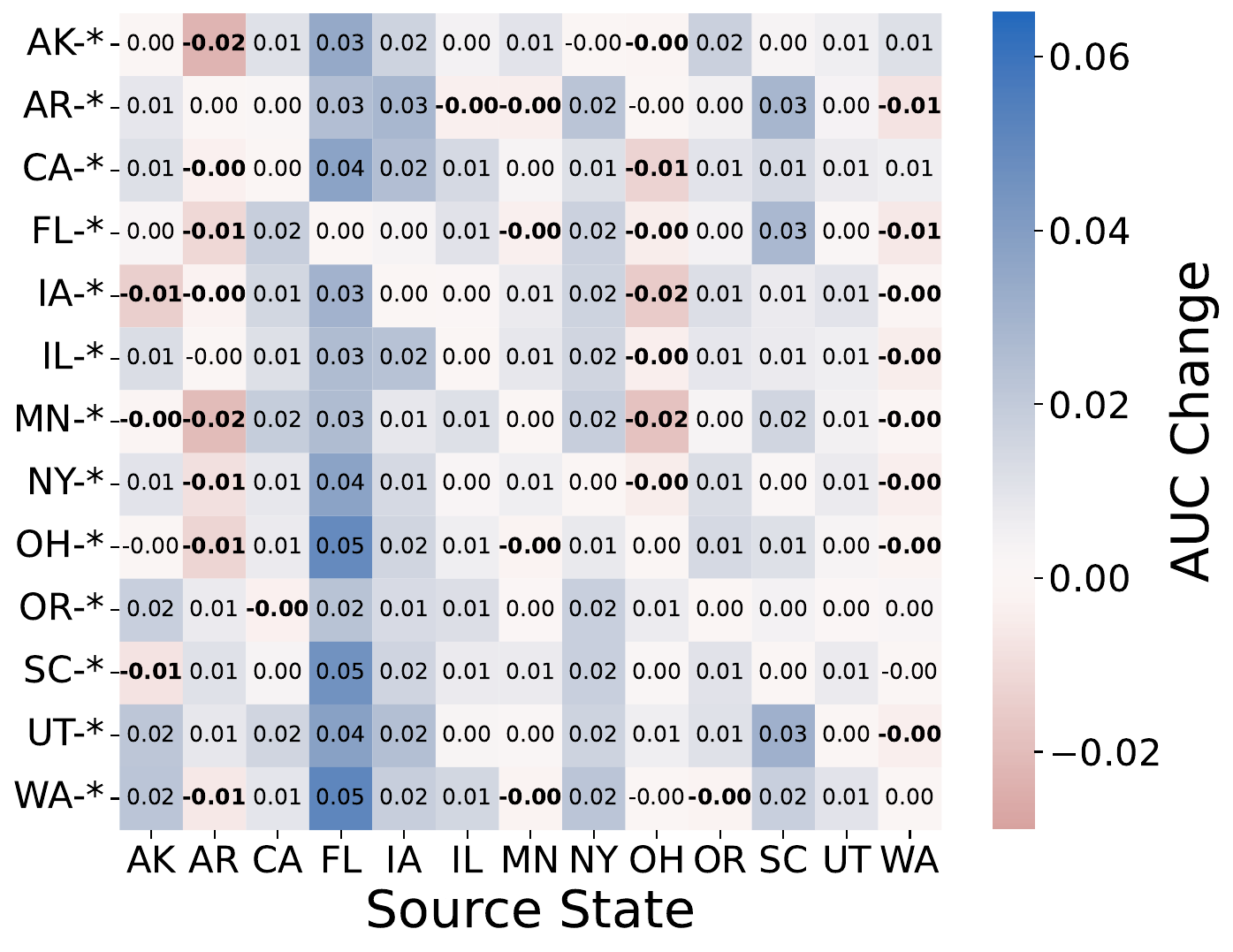}
    \caption{In Folktables~\cite{Ding2021RetiringAN}, combining with random states leads to worse income prediction in 47 out of 50 states.}
    \label{fig:enter-label}
\end{figure*}

\begin{figure*}[t]
    \centering
    \includegraphics[width=0.98\linewidth]{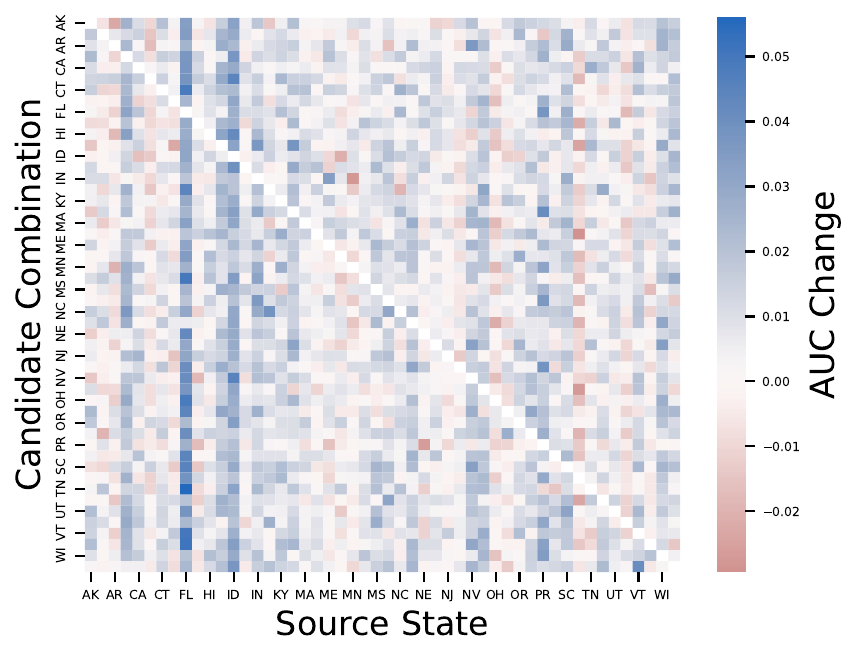}
    \caption{In Folktables~\cite{Ding2021RetiringAN}, combining with random states leads to worse income prediction in 47 out of 50 states.}
    \label{fig:enter-label}
\end{figure*}

\section{Additional Related Work}
\subsection{Data Valuation and Pricing}
The question of how to assess the impact and therefore worth of data has been well studied. Data valuation as a field seeks to quantify the contribution of individual data points or datasets to model performance. Shapley value-based approaches provide theoretically grounded valuations but scale poorly to large datasets~\cite{ghorbani2019data}. More efficient methods include influence functions~\cite{koh2017understanding} and leave-one-out testing~\cite{cook1979influential}. Recent work has extended these concepts to dataset-level valuation~\cite{jia2019towards}. As practitioners create larger datasets, the emergence of data marketplaces has sparked interest in data pricing mechanisms~\cite{kumar2020marketplace}. Query-based pricing~\cite{koutris2015query} and outcome-driven valuations~\cite{radic2024pricing} aim to balance seller compensation with buyer utility. While these approaches inform fair data exchange, they typically assume direct access to data, unlike our privacy-preserving method.
\subsection{Alternative Approaches to Data Sharing}
Recent work has explored several approaches to mitigate data sharing constraints while maintaining model performance. %We discuss two primary directions: synthetic data generation and transfer learning from public pretraining.

\paragraph{Public pre-training and private fine-tuning}

Transfer learning via public pretraining has become increasingly popular for domains with limited private data access. BioBERT~\cite{lee2020biobert} and ClinicalBERT~\cite{alsentzer2019publicly} demonstrated that pretraining on PubMed abstracts and clinical notes can improve performance on downstream medical tasks. Similar approaches have emerged in other regulated domains, including FinBERT~\cite{liu2021finbert} for financial applications.
However, the effectiveness of transfer learning depends heavily on domain alignment. One study showed that continued pretraining on domain-specific data significantly outperforms generic pretraining when domains differ substantially~\cite{gururangan2020don}. This presents challenges for highly specialized fields where public data may not capture domain-specific patterns~\cite{le2022language}.
Recent work has explored methods to quantify and optimize domain adaptation. Adaptive pretraining strategies~\cite{howard2018universal} and domain-specific vocabulary augmentation~\cite{gururangan2020don} have shown promise in bridging domain gaps. However, these approaches still require substantial compute and may not fully capture specialized domain knowledge present in private datasets.

\paragraph{Secure and Confidential Computation}
This requires an encoding scheme $\mathrm{Enc}(\cdot)$ that satisfies the homomorphic property: $\mathrm{Enc}(A)\circ \mathrm{Enc}(B) = \mathrm{Enc} (A\circ B)$, where $A$ and $B$ represent data held by two parties. The inverse function $\mathrm{Enc}^{-1}(\cdot)$ must exist to decode the final output: $\mathrm{Enc}^{-1}(A\circ B ) = A\circ B $. Considering an ``honest-but-curious'' threat model, where parties aim to jointly compute on privately-held data, two main approaches emerge.

% Input privacy guarantees that no intermediate information reveals original data.  This guarantee is derived from cryptographic primitives. Specifically in machine learning, preserving input-privacy allows for shared computation on hidden data, because the setup is cryptographically secure against data leakage.

% Though our method is designed for a two-party threat model, the underlying technique of secure data addition appraisal applies to semi-honest setup of multiple parties.

\paragraph{Fully Homomorphic Encryption (FHE)} FHE enables arbitrary additions and multiplications on encrypted inputs. While it represents the gold standard for encrypted computation, adapting it to modern machine learning is challenging due to computational constraints from growing ciphertext size. FHE implementations typically use lattice-based schemes requiring periodic ``bootstrapping'' (key refreshing and noise reduction through re-encryption) via methods like the CKKS scheme~\cite{cheon2017homomorphic}. This introduces cryptographic parameters that non-experts struggle to configure effectively.

\paragraph{Differential Privacy} 

Differential privacy (DP)~\cite{dwork2006differential} offers formal privacy guarantees for sharing data and training machine learning models. While DP mechanisms can protect individual privacy when releasing model outputs or aggregated statistics, they face significant limitations for inter-organizational data sharing. The primary challenge is that DP operates on already-pooled data, but organizations are often unwilling to share their raw data in the first place~\cite{dwork2014algorithmic}.
Even when organizations are willing to share data, the privacy guarantees of DP come at a substantial cost to utility, particularly in machine learning applications. DP-SGD, the standard approach for training deep neural networks with differential privacy, significantly degrades model performance compared to non-private training~\cite{abadi2016deep}. This performance impact is especially pronounced in data-constrained settings, where recent work has shown that large models rely heavily on memorization of rare examples that DP mechanisms tend to obscure~\cite{feldman2020neural}.
The privacy-utility trade-off becomes even more challenging when dealing with high-dimensional data or complex learning tasks. Studies have demonstrated that achieving meaningful privacy guarantees while maintaining acceptable model performance requires prohibitively large datasets~\cite{bagdasaryan2020backdoor}. This limitation is particularly problematic in specialized domains like healthcare, where data is inherently limited and performance requirements are stringent~\cite{geyer2017differentially}.
Recent work has attempted to improve the privacy-utility trade-off through advanced composition theorems and adaptive privacy budget allocation~\cite{papernot2021tempered}. However, these approaches still struggle to match the performance of non-private training, especially when working with modern deep learning architectures~\cite{tramer2020differentially}. While differential privacy offers important theoretical guarantees, our work focuses on the practical challenge of enabling data owners to evaluate potential partnerships before sharing any data, addressing a key barrier to collaboration that DP alone cannot solve.
\label{app:extra_related}
\end{document}